\documentclass[sigconf,10pt]{acmart}

\usepackage{tikz}
\usepackage{amsmath}
\usepackage{subfigure}
\usepackage{url}
\usepackage{booktabs}
\usepackage{xcolor}
\usepackage[linesnumbered,ruled,vlined]{algorithm2e}

\usepackage{amsmath}
\newcommand{\myeqref}[1]{Eq.\eqref{#1}}
\newcommand{\myref}[1]{Algorithm \ref{#1}}
\SetCommentSty{mycommfont}
\begin{document}


\acmConference[Submitted to review for ACM MobiCom]{}{2024}{USA}

\settopmatter{printacmref=false} 
\renewcommand\footnotetextcopyrightpermission[1]{} 
\pagestyle{plain} 

\title{CloudEye: A New Paradigm of Video Analysis System for Mobile Visual Scenarios}
\author{
    Huan Cui\textsuperscript{1,2}, Qing Li\textsuperscript{3,*}, Hanling Wang\textsuperscript{1}, Yong Jiang\textsuperscript{1} \\
    \textsuperscript{1}Tsinghua University \\
    \textsuperscript{2}Peking University \\
    \textsuperscript{3}Peng Cheng Laboratory
}


\begin{abstract}
  Mobile deep vision systems play a vital role in numerous scenarios. However, deep learning applications in mobile vision scenarios face problems such as tight computing resources. With the development of edge computing, the architecture of edge clouds has mitigated some of the issues related to limited computing resources. However, it has introduced increased latency. To address these challenges, we designed CloudEye which consists of \textit{Fast Inference Module}, \textit{Feature Mining Module} and \textit{Quality Encode Module}. CloudEye is a real-time, efficient mobile visual perception system that leverages content information mining on edge servers in a mobile vision system environment equipped with edge servers and coordinated with cloud servers. Proven by sufficient experiments, we develop a prototype system that reduces network bandwidth usage by 69.50$\%$, increases inference speed by 24.55$\%$, and improves detection accuracy by 67.30$\%$.
\end{abstract}



\maketitle

\section{Introduction}
At present, mobile deep vision systems play a vital role in numerous scenarios, such as drones, autonomous vehicles, and intelligent monitoring systems \cite{hu2023planning, peng2023skynet}. The realization of the future metaverse will also greatly rely on mobile deep vision technologies. Due to availability of massive data and advancements in deep learning model performance\cite{suprem2020odin, canel2019scaling, kang2018blazeit}, deep learning-based mobile vision systems are transitioning from being merely usable to highly effective. However, there are various challenges in this process. Deep learning models typically have a vast number of parameters, and computation-intensive deep vision systems are constrained by limited processing power on mobile devices\cite{khani2023recl, wen2023adaptivenet, padmanabhan2023gemel}. Moreover, mobile deep vision systems are often latency-sensitive, such as in autonomous driving and augmented reality applications, where system latency significantly impacts performance, requiring rapid environmental perception and real-time processing.

To address these challenges, researchers have proposed several strategies to enhance the performance of mobile vision systems. Offloading computation to cloud servers with powerful computational resources is suggested to ensure the accuracy of vision tasks\cite{zhang2023crossvision, li2020reducto, zhang2017live, kang2017noscope, wu2021pecam}. However, this approach heavily dependends on high-speed, low-latency networks \cite{meng2023enabling, cangialosi2022privid}. Fluctuating or slow networks may have disastrous effects on system performance in scenarios like autonomous driving. Additionally, the environments sensed by mobile vision systems are often in the form of videos, which exhibit strong spatio-temporal correlations. Video frames in historical sequences contain redundant information\cite{li2022livenet}. Objects in video frames that are temporally close also tend to be spatially close. This is determined by the kinematics of macroscopic objects. Offloading computation to cloud servers in the form of frames does not utilize this spatio-temporal correlation\cite{jain2018rexcam}. On the other hand, edge servers are designed to ensure real-time processing\cite{xiao2021towards,yang2022edgeduet}, allowing model inference to occur on mobile devices\cite{huang2022real}. However, high-resolution videos with richer information better serve visual tasks in practical applications\cite{zhang2021elf}, and the limited hardware capabilities of edge servers cannot independently support the operation of large models for high-resolution image inference \cite{luo2021resource, zhou2019edge}.

To this end, combining the high-precision results generated by cloud server computational resources with the low-latency response characteristics of mobile edge servers can balance system accuracy and real-time performance\cite{du2022accmpeg,zhang2022understanding,hsieh2018focus}. Due to the strong spatio-temporal correlations in videos, fully exploiting the information in historical inference results of edge servers can significantly enhance the inference effect on the current frame. Some researchers have used the calculation of reference frame and current frame feature points and their matching to utilize historical information\cite{chen2015glimpse, yang2022flexpatch}, but video scenes are subject to subtle changes in lighting and shadows. Traditional handcrafted feature points based on pixels are susceptible to minor disturbances in videos\cite{wojke2017simple}. Through experiments, we discover that although weaker models on mobile edge servers perform poorly in directly regressing inference results, they have a strong ability to extract features and represent information in video frames\cite{du2023strongsort, hou2022neulens}. These models maintain considerable stability when representing identical content information across different frames. In other words, weaker models on mobile edge servers have considerable potential. Attention mechanisms are also well-suited for the spatio-temporal correlations in videos\cite{liu2019edge, vaswani2017attention}, and performing inference around regions of interest containing key content in the current frame optimizes model execution.

There are several challenges to overcome. First, system resources are fragmented due to constraints such as the bandwidth with cloud servers and the processing power of mobile edge servers. A dynamic system resource allocation scheme should be developed to ensure reasonable task offloading and prevent any part from becoming a resource bottleneck\cite{ouyang2022cosmo}. Second, designing an attention mechanism suitable for various visual tasks is crucial. A robust mechanism should be developed for predicting ROI and optimizing the inference process around them. Finally, fully utilizing the results from cloud servers and the potential of mobile edge servers requires designing an efficient, unified system framework.

To solve these problems, we designed \textbf{CloudEye}, a real-time, efficient mobile visual perception system that leverages content information mining on edge servers in a mobile vision system environment equipped with edge servers and coordinated with cloud servers. Under the guidance of high-precision information from cloud servers, CloudEye mines content information on edge servers to achieve real-time, efficient mobile visual perception. We implement this system through three approaches. First, we propose an attention mechanism that, guided by discrete cloud server inference results, refers to historical information to mine the content to be perceived in the current frame and utilizes edge model features for content information mining. Second, we determine regions of interest based on the mined content distribution and accelerate the inference process for high-resolution images accordingly. Lastly, CloudEye adopts a lightweight approach, dynamically filtering and differentially compressing video frames transmitted to cloud servers based on system resources such as bandwidth to ensure information density in regions of interest while compressing background information, maximizing resource efficiency.

The main contributions of this paper are as follows:

\begin{itemize}
    \item We propose, for the first time, a system that fully mines information from historical inference results on edge servers under cloud server guidance to optimize mobile visual tasks. This system fully exploits the content feature information extracted by weaker models on edge servers. 
    \item We introduce an attention mechanism that includes ROI prediction, dynamic video frame compression based on regions of interest, and an accelerated model inference process. Additionally, we design a compensation scheme to improve the stability of the system.
    \item We develop a prototype system that reduces network bandwidth usage by 69.50\%, increases inference speed by 24.55\%, and improves detection accuracy by 67.30\%.
\end{itemize}


\section{MOTIVATION and CHALLENGES}

\subsection{Mobile Visual Scenarios}

The objective of this study is to investigate deep learning models applied to mobile visual scenarios. In these scenarios, systems often execute tasks such as object tracking, human pose estimation, image segmentation, and other vision-based tasks. These tasks typically involve processing video data, using deep learning models to extract image features and generate proposals containing target objects. Subsequently, the system classifies or recognizes the targets within the proposals. With the widespread use of smartphones and virtual reality devices, mobile vision has become an integral part of people's daily lives.

\subsection{Limitations of Existing Solutions}

Deep learning visual models generally have a large number of parameters\cite{jiang2018chameleon}. Due to the inherent requirements of mobile visual scenarios, mobile devices are often lightweight and have limited computational power\cite{yi2020eagleeye}. This makes it challenging to run deep learning visual models at high frame rates, hindering real-time performance. Moreover, mobile visual scenarios increasingly demand higher video resolution, leading to substantially increased data transmission volumes and significant bandwidth consumption during network transmission or offloading\cite{yeo2022neuroscaler}.

Researchers have proposed various solutions to address these challenges and improve the performance of mobile visual systems. 

To utilize the spatio-temporal correlation in video, some have proposed using handcrafted feature points to match reference frames with the current frame \cite{cvivsic2018soft, lindeberg2012scale}. However, subtle changes in illumination and shadows can impact the performance of pixel-based handcrafted features. Others have suggested tracking based on results from cloud servers, but the characteristics of handcrafted features limit processing speed and accuracy\cite{chen2015glimpse}.

Some suggest offloading computation to powerful cloud servers to ensure task accuracy \cite{du2020server}, but this relies heavily on high-speed, low-latency networks \cite{zhang2017live}. In scenarios such as autonomous driving, fluctuating or slow networks can have catastrophic effects on system performance. Additionally, since mobile visual systems usually perceive their environments through video, there is strong spatio-temporal correlation among frames \cite{li2020reducto}, which is not exploited when offloading computation to cloud servers frame by frame \cite{zhang2021elf, jiang2021flexible}.

On the other hand, edge servers have been designed to ensure real-time performance by allowing model inference to take place on mobile devices. An edge server is a micro-server with customized CPU, GPU and operating system. It can be deployed mobilely and has certain computing power and deep learning model inference capabilities. However, models with excellent performance in real-world vision tasks often have large number of parameters, and the hardware resource of edge servers is weak and cannot support the operation of large models \cite{mullapudi2019online, bhardwaj2022ekya}. 

To optimize the workload, some researchers have proposed hybrid edge-cloud architectures for task offloading to balance computing resources. In this approach, edge servers perform inference during cloud server inference gaps, but the weaker edge servers yield lower accuracy for inferred frames \cite{hu2019dynamic}, and transmitting frames to cloud servers consumes considerable bandwidth \cite{zhang2022batch, zhang2022casva}. In scenarios such as autonomous driving in remote areas, or in places where internet speed is only a few tens of KB/s \cite{zhang2015design}, these limitations are particularly problematic.

Existing models typically perform inference on all regions of video frames \cite{ren2015faster, huang2018yolo}, assuming that targets may appear anywhere in the frame. This approach does not utilize the reference information from historical frames and contributes to increased inference latency and computational resource consumption.

To address the limitations of existing work, we propose CloudEye.

\subsection{Design Challenges}

In CloudEye, we aim to enable edge servers to perform high frame-rate inference on video frames. However, edge servers have limited computing resources, and many high-precision models with large parameter volumes cannot run on them. For example, most mainstream object detection models are quite large and resource-intensive, making them unsuitable for deployment on edge servers. Therefore, the design of CloudEye needs to address the following challenges:

\textbf{Model optimization:} Develop lightweight deep learning models for mobile visual tasks that are capable of running on edge servers with limited computing resources. This requires designing models with fewer parameters and reduced complexity while maintaining acceptable accuracy.

\textbf{Spatio-temporal correlation exploitation:} Leverage the spatio-temporal correlation in video frames to optimize the computation and data transmission process. This involves tracking target objects across frames and using historical information to guide the inference process, reducing the need for full-frame inference and lowering computational resource consumption.

\textbf{Adaptive offloading strategy:} Design an intelligent offloading strategy that can dynamically allocate computing tasks between edge servers and cloud servers based on the current network conditions and the computational capabilities of edge servers. This will help to balance the workload, reduce latency, and save bandwidth.

\textbf{Scalability and generality:} Ensure that the proposed CloudEye framework is scalable and can be applied to various mobile visual tasks, such as object tracking, human pose estimation, and image segmentation. This includes designing modular components that can be easily adapted or replaced to support different tasks and scenarios.

\section{OVERVIEW AND DESIGN GUIDELINES}

\begin{figure}[h]
  \centering
  \includegraphics[width=0.97\linewidth]{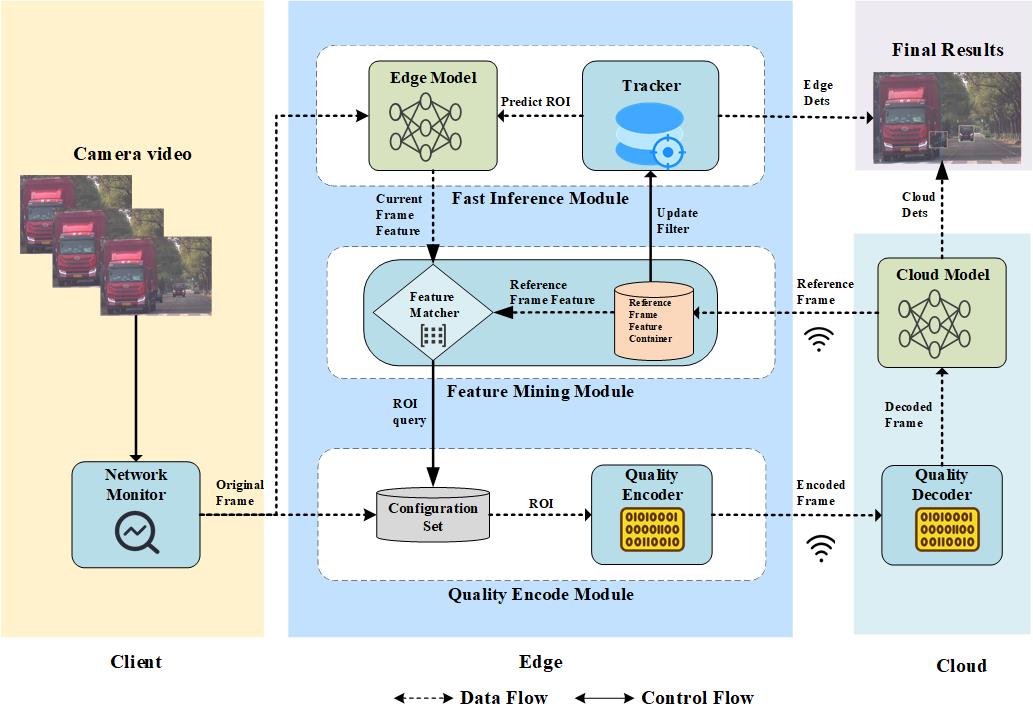}
  \caption{System Architecture}
  \label{fig:architecture}
\end{figure}

In summary, as depicted in Figure~\ref{fig:architecture}, we propose the following design for CloudEye:

1) Develop a \textit{Fast Inference Module} that adapts the edge model's structure to accommodate the limited computing resources available on edge servers;

2) Implement a \textit{Feature Mining Module}, in which the edge model on the edge server infers video frames and uploads key frames to the cloud server for further processing. Utilizing deep learning features extracted by the edge model, a lightweight and efficient matching process is conducted with the cloud server's inference results. This approach transforms the deep learning model's object detection regression problem into a similarity comparison problem for locating matching regions;

3) Design a \textit{Quality Encode Module} that dynamically applies differentiated quality encoding to video frames transmitted to the cloud server. This technique ensures the accuracy of the cloud server's key frame inference while simultaneously reducing bandwidth consumption.

4) According to the characteristics of the mobile visual scenarios, CloudEye is designed in modules and is implemented on mobile vision devices, so that it can be deployed on existing mobile vision devices at low cost.

To ensure real-time system performance and enable instantaneous edge server inference, the \textit{Fast Inference Module} exploits the spatio-temporal correlations in videos, predicts image locations based on historical information, and performs inference in regions of interest. Experimental results demonstrate that the \textit{Fast Inference Module} significantly accelerates the inference speed of content-aware deep learning visual models, such as object detection models, without sacrificing accuracy. This greatly reduces the computational and energy consumption of edge servers.

For the feature extraction module, we employ a tracker based on the Kalman filter \cite{bewley2016simple} to track targets. Both edge and cloud servers infer the current frame to obtain observations, filtering the content distribution in accordance with motion laws. With the computational power of our robust cloud server and the high-precision model deployed, the cloud server's results are highly accurate. However, due to bandwidth limitations, we cannot guarantee the system's real-time performance by relying on a predetermined set of rules for sending frames to the cloud server. Edge servers continually infer video frames, ensuring real-time system performance. Owing to the edge server's limited computing power, the edge model cannot produce complete and accurate results. Instead, the cloud server's results offer precise guidance for content information distribution. For targets the edge model cannot recognize or exhibits significant deviation, the feature extraction module combines the cloud server's results and the edge model's extracted deep visual features from video frames. Ultimately, the precise distribution regions of target content in the current video frame are determined by synthesizing the edge server's inferred and extracted results, and the tracker is updated accordingly.

At the same time, the system must upload frames to the cloud server following specific rules to obtain high-precision model results, while it often operates in bandwidth-constrained and complex scenarios. Thus, it is essential to reduce data transmission volume with the cloud server while maintaining system accuracy. We find that the current frame can be significantly divided into background and regions of interest based on the distribution of content in historical frames based on video spatio-temporal correlations. The \textit{Quality Encode Module} applies dynamic compression to video frames, guided by regions of interest, ensuring information density in areas of interest while reducing background entropy. Empirical evidence reveals that the robust region-of-interest segmentation and compression algorithms we propose do not affect the accuracy of the cloud server's inference results, while substantially reducing data transmission volume.

In the following sections, we will discuss the implementation of each module in detail.


\section{Fast Inference Module}
Existing mobile vision applications primarily employ models tailored for image processing, which consume significant computational power. In practical deployments, however, mobile vision systems operate by processing videos. Although videos are parsed into individual frames for processing, they exhibit strong spatio-temporal correlations. Treating video frames as independent images can neglect the potential benefits of leveraging these correlations to enhance video analysis and introduce redundant computations. Therefore, the algorithm should utilize spatio-temporal correlations while eliminating redundant calculations to reduce computational overhead.

\subsection{Module Implementation}
\begin{figure}[h]
	\centering  
	\subfigbottomskip=2pt 
	\subfigcapskip=-5pt 
	\subfigure[Original model proposals]{
		\includegraphics[width=0.48\linewidth]{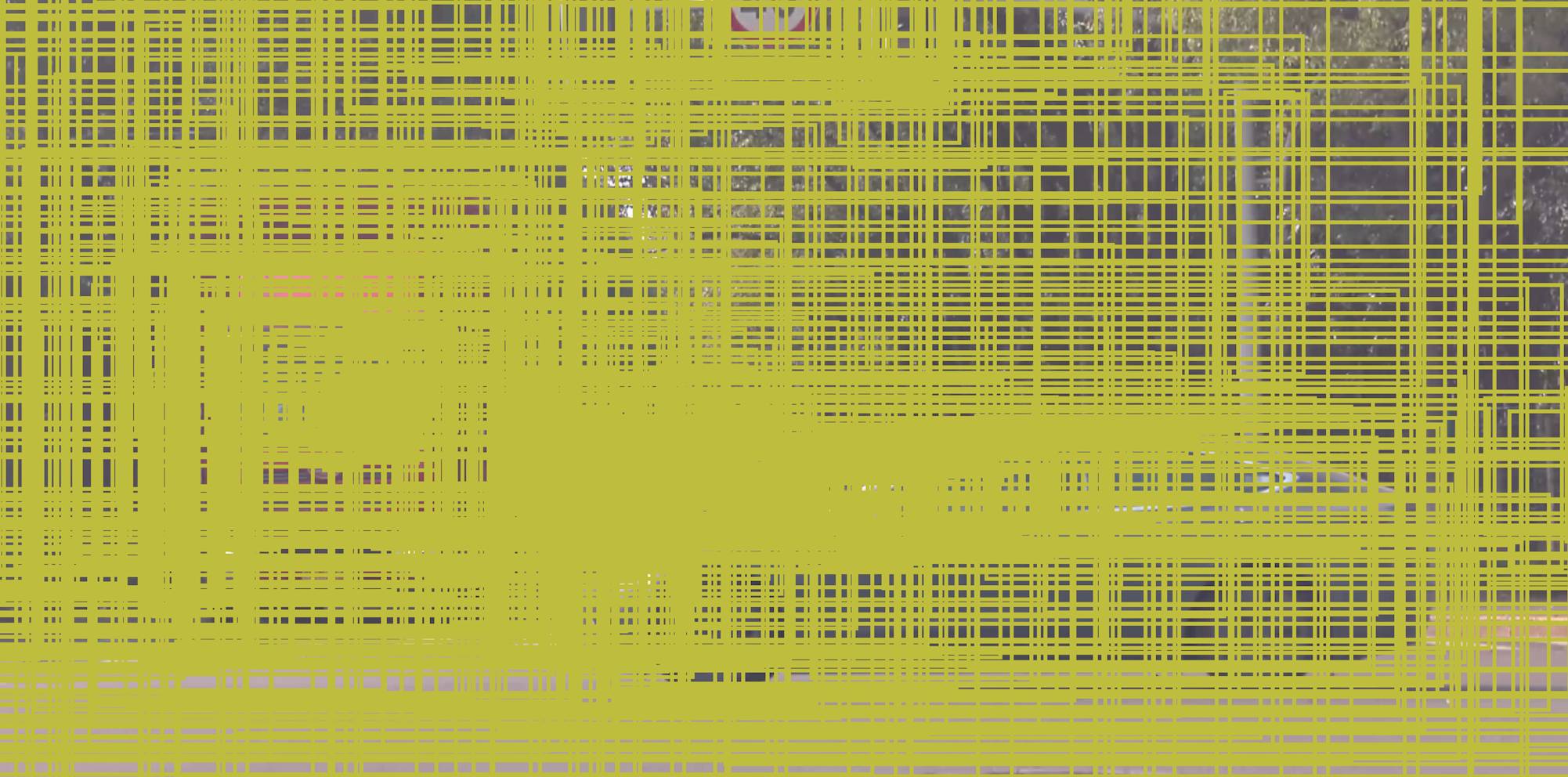}}
	\subfigure[CloudEye proposals]{
		\includegraphics[width=0.48\linewidth]{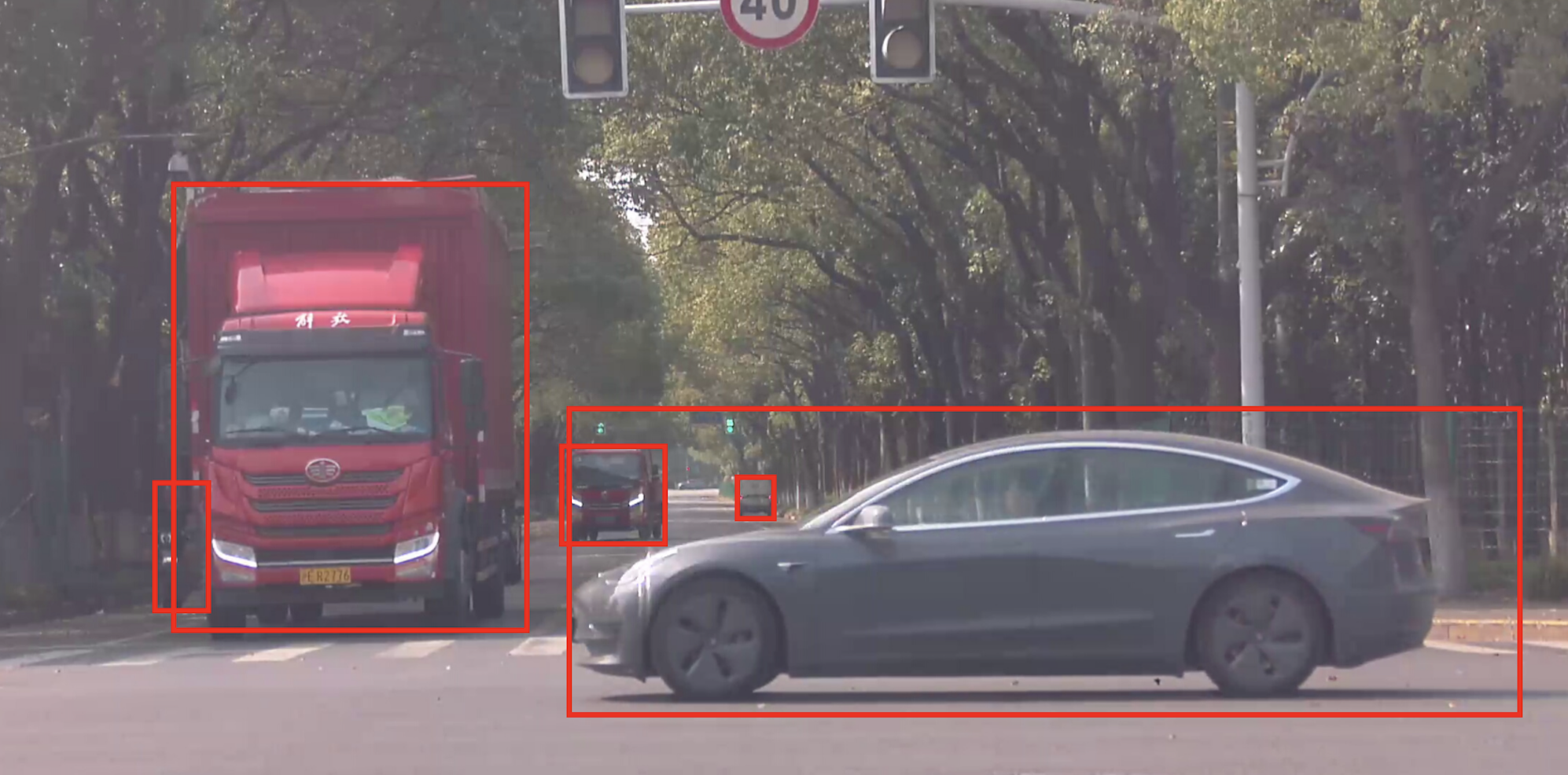}}
        \caption{Proposals Comparison: the subgraph (a) shows all the proposals generated by the original model, and the red boxes in the subgraph (b) are the prior proposals provided by CloudEye's \textit{Fast inference module}.}
        \label{fig:pro1}
\end{figure}

To exploit the content of previous frames in a temporal sequence, we use a tracker to assist in inferring the target distribution in the current frame. This tracker is implemented using a Kalman filter. For object detection models, these models generate proposals, which are essentially suggestions for the current target location, and subsequently refine their position estimates. Since these proposals do not consider the content distribution of previous frames, they may be excessive, consuming considerable computational power. Before processing video frames, we use the tracker to establish a kinematic prediction for the current target, serving as proposals for the current target. These proposals, together with the extracted current frame features $\mathbb{\hat F}$, are fed into the subsequent process to refine the correct position. This approach reduces computation and, because the tracker solves for the target's motion pattern, the proposals are reliable, decreasing the likelihood of false detection. As illustrated in the Figure~\ref{fig:pro1}, the \textit{Fast Inference Module} provides proposals that align with the target's precise position, whereas proposals generated by the model itself contain numerous false detection items.

On the other hand, due to the presence of noise, target detection exhibits discrepancies between the true and noisy distributions, and predictions also contain noise, both of which follow Gaussian distributions. Employing the Kalman filter can harmonize these discrepancies, yielding a more accurate Gaussian distribution for the probable target location, which is closer to the true distribution with reduced noise. This is known as the Kalman gain effect.

The algorithm workflow is as \myref{eq:fast}.

\begin{algorithm}
\DontPrintSemicolon
\caption{Fast Inference}
\label{eq:fast}
\SetAlgoLined
\KwIn{Tracking module $KF$, detection boxes in last frame $\textbf{O}=\{O_1, O_2, ..., O_n\}$}
\KwOut{Current frame boxes $\hat{\textbf{O}}$ and updated $KF$}

\tcc{Step-1. Tracking module predicts proposals}
$\textbf{O}_{pre} \leftarrow \operatorname{KF.predict}(\textbf{O})$ \\
\tcc{Step-2. Edge model fast inference}
$\mathbb{\hat F} \leftarrow \operatorname{EdgeModel.Extract}(Img)$ \\
$\textbf{O}_{det} \leftarrow \operatorname{EdgeModel.Regression}(\textbf{O}_{pre},\mathbb{\hat F})$ \\
\tcc{Step-3. Filter and update}
$\hat{\textbf{O}} \leftarrow \operatorname{KF.update}(\textbf{O}_{det},\textbf{O}_{pre})$ 

\end{algorithm}

Thus, the \textit{Fast Inference Module} assists us in leveraging spatio-temporal correlations for more accurate detection, reducing computational load, and utilizing the Kalman gain to achieve greater target position accuracy. The design of the \textit{Fast Inference Module} allows it to be applied to various video analysis systems based on object detection models.

\subsection{Appendix Case}

In the Fast Inference Mode, proposals are derived from targets that have previously appeared within the field of view in historical frames. The system will conduct a full model inference on the video frame under two circumstances. The first is when there is a significant change in video pixel values. When the difference between video pixel frames exceeds a certain threshold, the scene is likely to undergo a substantial transformation. In this case, the mobile device may have moved to an entirely new environment, with new targets appearing in the video. Consequently, it is necessary to perform a full model inference on the video frame. The second circumstance arises when the bandwidth is insufficient, and the system has not sent frames to the cloud server or received results from it for an extended period. In this situation, a full model inference should also be performed on the current video frame.

\section{Feature Mining Module}
Our aim is to equip edge servers with models capable of performing high frame-rate inference on video frames, necessitating CPU computation and GPU inference. Hence, the feature extraction algorithm should be lightweight. Furthermore, the algorithm should effectively refer to historical frames to extract the target position in the current frame and ensure that the extracted target trajectory adheres to the laws of motion.

\subsection{Module Implementation}

This module has two main stages as follows.

\subsubsection{Get Cloud Server Results}

The \textit{Feature Mining Module} will utilize the high-precision inference results from cloud servers on historical video frames to mine targets in the current frame. We determine which frames to upload to the cloud server based on the two principles. First, the system decides whether to send frames based on video frame pixel changes. As previously discussed, pixel changes in video frames signify content changes, and key frames with dramatic pixel changes are generally uploaded to the cloud. Second, whether to send frames to the cloud server is determined based on the current bandwidth between the edge server and the cloud server. We have set up a queue for frames awaiting transmission, and when the bandwidth permits, we arrange the frames in chronological order in the queue and use a certain probability to decide whether to send the frame at the head of the queue. This approach ensures key frames with pixel changes have higher priority and are more likely to preemptively be sent to the cloud server, preventing regular frames from blocking future key frames.

\subsubsection{Feature Extraction at the Edge Server}

We store the finely-tuned model inference frames, obtained from the cloud server, as reference frames, and cache the local frames inferred by the edge server between those reference frames and the current frame. When the time interval between the current frame and the cloud server's inferred frame is brief, we directly employ the cloud server's inferred frame as the reference frame for extracting features. Conversely, when the time interval between the cloud server's inferred frame and the current frame is extensive, we rely on the cached local frames to extract features up to the current frame. The threshold is determined by the average speed of the target's motion, which corresponds to the average dissimilarity in content.

Having established the reference frame and the chosen extraction method, for the target $O_{i}$ to be extracted in the reference frame, the system selects the appropriate feature layer based on the scale size of the target $O_{i}$. As features provide a robust representation of the target, the reference frame generally encompasses regions with high confidence and accurate positions through the cloud-based high-precision model inference. The features of the target area extracted in the current frame should closely resemble those of the target in the reference frame. Our optimization objective is to minimize the discrepancy between the features of the target area $O_{i}$ in the reference frame, denoted as $\mathbb{F}{O{i}}$, and the features of the target area $\hat O_{i}$ extracted in the current frame, denoted as $\mathbb{\hat F}{\hat O{i}}$, which is expressed as $\mathcal{L}\left(\mathbb{F}{O{i}}, \mathbb{\hat F}{\hat O{i}}\right)$.

\begin{equation}
\begin{aligned}
&\min _{f} \mathcal{L}\left(\mathbb{F}_{O_{i}}, \mathbb{\hat F}_{\hat O_{i}}\right) = \min \sum_{x, y} \| F_{(x, y)}, \hat F_{f(x, y)} \| \\
&\text { s.t. } f(x, y) = \textbf{S} \cdot[x, y]^{\top}+\textbf{T}, \\
&\textbf{S}=\left[\begin{array}{cc}
s_{x} & 0 \\
0 & s_{y}
\end{array}\right], 
\textbf{T}=[t_{x}, t_{y}]^{\top} 
\end{aligned}
\end{equation} 

\noindent the coordinates $(x, y)$ belong to the target $O_{i}$, representing the pixel coordinates of the target within the image frame, as illustrated in Figure 3. The column feature vector $F_{(x, y)}$ is derived by convolving the target pixel position, serving as the receptive field center, with the selected feature layer appropriate for the target's scale. The function $f$ denotes the transformation from $O_{i}$ to $\hat O_{i}$, while $\textbf{S}$ denotes the scaling matrix and $\textbf{T}$ represents the translation matrix. The features corresponding to $O_{i}$ and $\hat O_{i}$ are denoted by $\mathcal{L}$, where the Mahalanobis distance is employed as a measure of similarity between the features.

Due to the high dimensionality of the feature vector (e.g., ResNet18 can extract 256-dimensional feature vectors) and the large number of feature vectors corresponding to target area pixels in the feature layer, computing the similarity of feature vectors poses a computationally intricate challenge. Nevertheless, as evidenced by prior research, the deep neural networks' features, with their expansive receptive fields and robust representation capabilities, effectively encapsulate a significant portion of the target's content information within the feature vectors corresponding to the target areas. Hence, by selectively sampling a few feature vectors at reasonable intervals from these regions, we can obtain a set of feature vectors that comprehensively represent the entire target content information, enabling us to search for them within the current frame. Concerning the search area, the system automatically calculates the ROI $S_{i}$ for the target search. This region is defined by the predicted position $O_{pre,i}$ obtained from the Kalman filter and the target's position $O_{i}$ in the reference frame. That is,

\begin{equation}
\begin{aligned}
&S_{i} = \operatorname{Combine}(O_{pre,i},O_{i}) \\
&e.g.\quad A_{max,S_{i}}=Max(A_{max,O_{pre,i}},A_{max,O_{i}})+l_{padding} \\
&\qquad \quad A_{min,S_{i}}=Min(A_{min,O_{pre,i}},A_{min,O_{i}})-l_{padding}
\end{aligned}
\end{equation}

\noindent where $\textbf{A}_{max/min}$ denotes the extreme values of the target pixel area coordinates, and $l_{padding}$ is a hyperparameter representing the search area's expansion magnitude.

\begin{figure}[h]
    \centering  
    \subfigbottomskip=2pt 
    \subfigcapskip=-5pt 
    \subfigure[Ref frame]{
        \includegraphics[width=0.48\linewidth]{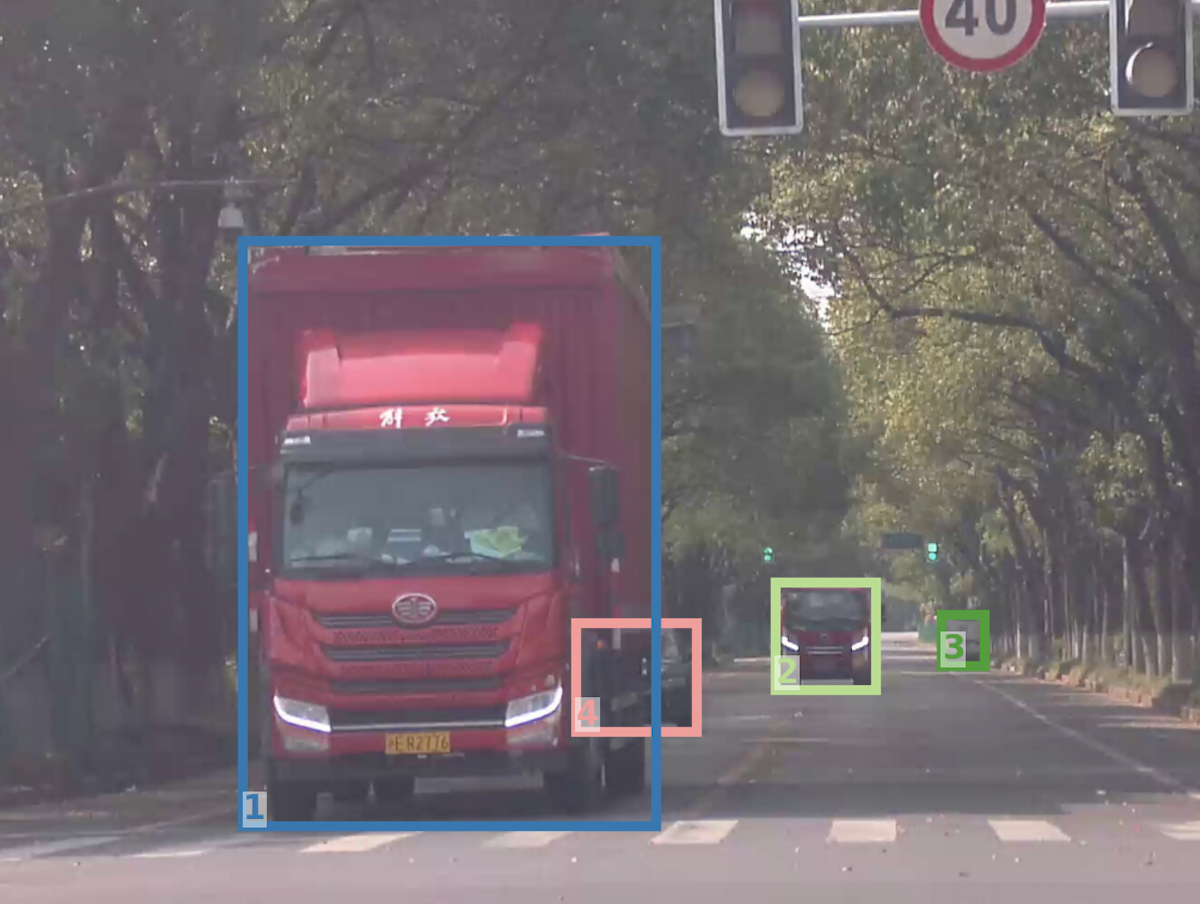}}
    \subfigure[Current frame]{
        \includegraphics[width=0.48\linewidth]{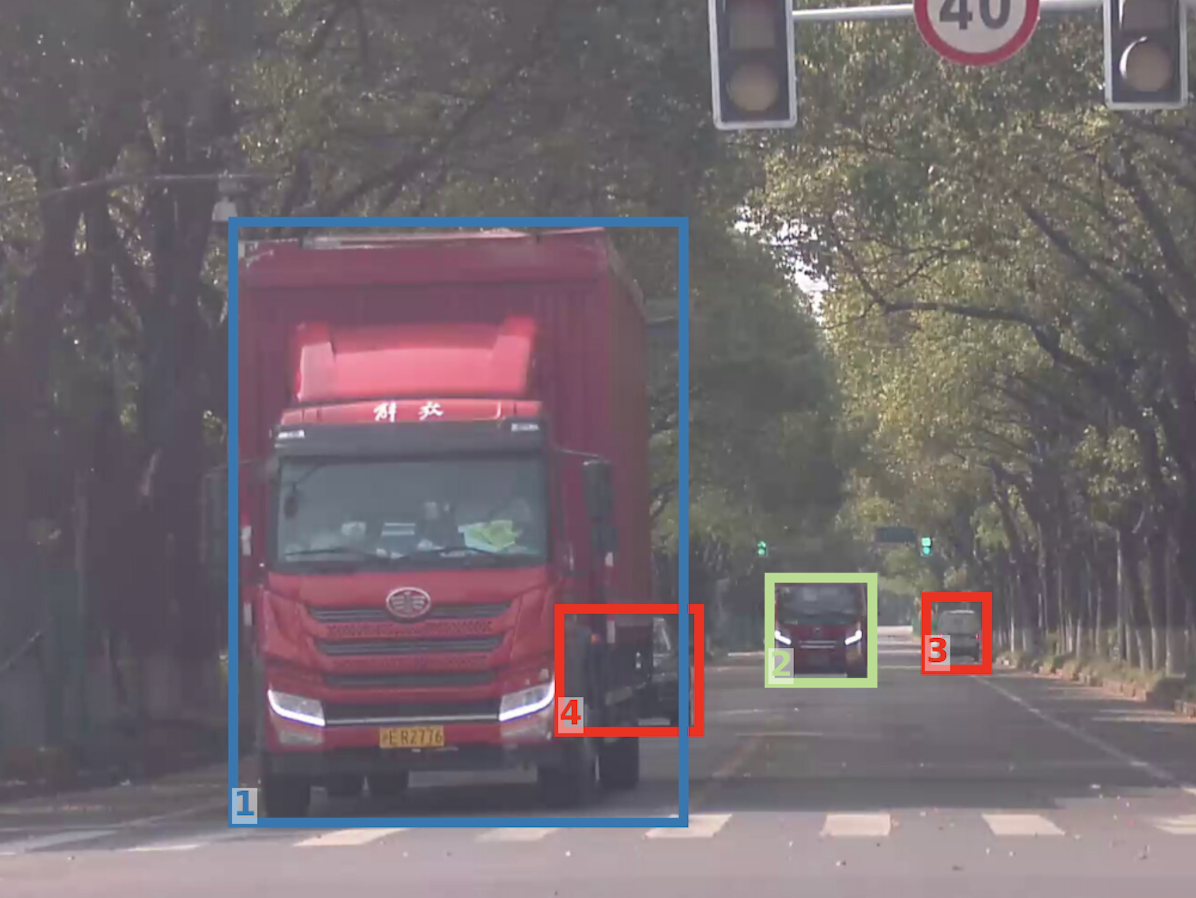}}
    \caption{Feature Mining Module Effect: in subgraph (a), the colored boxes are the targets of the reference frame. In subgraph (b), the colored boxes except the red boxes are objects detected by the edge model, while the red boxes are additional objects detected by CloudEye's \textit{Feature Mining Module}.}
    \label{fig:mine}
\end{figure}

Our subsequent experiments also verified the effectiveness of this strategy. As shown in Figure \ref{fig:mine}, we matched the sampled feature vectors, obtaining the sampling point set $\textbf{P}=\left\{p_{i}\right\}$, $i \in[0,1, \ldots, 4]$ in $O_{i}$, where $p_{0}$ is the geometric center of $O_{i}$ and $p_{1-4}$ are the geometric centers of the four subregions that divide $O_{i}$. First, the feature vector $F_{p_{0}}$ corresponding to $p_{0}$ in the reference frame is searched and compared for similarity within the range of $S_{i}$ in the feature layer selected by $O_{i}$ in the current frame, and the position $\hat{p_{0}}$ with the highest similarity is obtained. Then, the feature vectors $F_{p_{1-4}}$ are searched and matched separately within the regions divided by $\hat{p_{0}}$ in $S_{i}$. By geometric relationships between $p_{1-4}$ and the entire target area box, the scale of the target box in the current frame is restored to obtain $\hat{O_{i}}$.

This approach not only reduces computational complexity but also achieves more robust results compared to manual feature point matching. By leveraging deep learning and effective feature mining techniques, the system can efficiently and accurately track and identify objects in video frames, even in edge computing environments with limited computational resources.

Furthermore, due to the overlapping receptive fields of convolution kernels in the feature layers, the feature vectors corresponding to pixel points with close target area coordinates exhibit relatively high similarity. However, under severe deformations such as object occlusions or sharp turns made by vehicles, the feature vectors may not match very similar vectors in the current frame. This can cause them to slide to nearby offset positions, resulting in unreasonable scaling and shape of the restored target area. To address this issue, when searching for a vector $f_2$ that matches the sampled vector $f_1$ in the current frame, if matching $f_2$ would lead to overall matching failure, a penalty strategy is applied to increase the similarity loss between $f_1$ and $f_2$ in the subsequent search. This helps the mined targets in the progressive search process gradually return to the correct position.

To achieve this, we utilize a loss function $L$ that is inversely proportional to the distance:

Furthermore, we set a threshold for the overall similarity of the sampled vector group. If the similarity loss exceeds this threshold, it indicates that the target is occluded by foreground objects. In such cases, we choose the point with higher matching similarity among the sampled feature vectors as the pivot and adopt a conservative strategy to mine the target area.

In the following \myref{eq:mine}, we will describe our algorithm process in detail. Subsequently, we proceed to update the tracker by incorporating both the mined targets and the detected targets. The \textit{Feature Mining Module} and the detection module collaborate to effectively track targets in the video stream. By leveraging the high-precision inference results from the cloud server and the lightweight edge server computation, the system can dynamically adjust the target tracking approach based on the current frame and historical frames. This adaptive combination ensures accurate target tracking, even in challenging scenarios involving occlusions or rapid target movements.


\begin{algorithm}
\DontPrintSemicolon
\caption{Feature Mining exploiting}
\label{eq:mine}
\SetAlgoLined
\KwIn{Object $O_{i}$ in the reference frame, object position $O_{pre,i}$ predicted by Kalman filter, features $\mathbb{F}$ of the reference frame on the selected feature layer and features $\mathbb{\hat F}$ of the current frame}
\KwOut{Discovered object $\hat O_{i}$, $Confidence(\hat O_{i})$}
\tcc{Step-1. Get feature search area}
  $S_{i} = \operatorname{Combine}(O_{pre,i},O_{i})$ 
 
\tcc{Step-2. Get $O_{pre,i}$'s points set $\textbf P^{\prime}$}
\For{$n$:1 to $Depth$}
{ 
   \For{$p_{i}$ in $\textbf{P}$}    
        { 
        $\hat{p_{i}} \leftarrow \operatorname{argmin}\left(\mathcal{L}\left(\mathbb{\hat F}, F_{p_{i}}\right)+L_{p_{i}}\right)$ \tcp*{$L$ is punished loss matrix when match $p_{i}$}
        }

   		\If{$\sum_{i} \mathcal{L}\left(\hat{F}_{\hat{p}_{i}}, F_{p_{i}}\right)<\varepsilon$}
    {
        break    \tcp*{searched the ideal area}
    }
    \ElseIf{$n<Depth-1$}
    {
    	Update $L$  with the inverse of the distance of the pixel point in search area from $p_{i}$        
    }
    \ElseIf{$n=Depth-1$}
    {
    	assign $\hat p_{0}$  to the center of $\hat O_{i}$ and make its scale equal to $O_{i}$    \tcp*{Conservative Strategy}
    }
   }
\tcc{Step-3. Get the size and position of $\hat O_{i}$}
$\hat O_{i,center} \leftarrow p^{\prime}_{0}$

$\hat O_{i,height} \leftarrow Distance(p^{\prime}_{1},p^{\prime}_{3}) + Distance(p^{\prime}_{2},p^{\prime}_{4})$

$\hat O_{i,width} \leftarrow Distance(p^{\prime}_{1},p^{\prime}_{2}) + Distance(p^{\prime}_{3},p^{\prime}_{4})$

\tcc{Step-4. Get the confidence of $\hat O_{i}$}

$Confidence(\hat O_{i}) \leftarrow \cfrac {\varsigma_{size(\hat O_{i})}} {\sigma(\mathcal{L}\left(\mathbb{F}_{O_{i}}, \mathbb{\hat F}_{\hat O_{i}}\right)) - \tau_{size(\hat O_{i})}}$

\KwRet $\hat O_{i},Confidence(\hat O_{i})$
  
        
\end{algorithm}

Our \textit{Feature Mining Module} does not impose any restrictions on the feature extraction model for deep learning. This means that it is designed to be plug-and-play, supporting any deep learning visual model to fulfill its task. For example, in the task of continuous segmentation of objects and areas in videos, our module provides prior guidance to predict and compare the ROI and the target position to be segmented for the image segmentation model.

\section{Quality Encode Module}
To facilitate feature mining, we employ an adaptive approach where a subset of frames is transmitted to the cloud for high-precision inference. In real-world scenarios, the distribution of perceived objects within the system is typically uneven. This means that the content information of foreground objects is often concentrated in a specific region of the video frame, while the remaining areas contain background information. Given that video analysis primarily focuses on foreground targets, traditional object detection methods that analyze and transmit the entire image are inefficient, redundant, and wasteful of computational resources. This is particularly problematic in bandwidth-limited mobile systems, as transmitting high-resolution images in their entirety incurs significant bandwidth consumption and latency.

To address these challenges, we need to dynamically plan the information density of different areas within the video frame that are transmitted to the cloud server. This planning should consider the varying importance of objects and the distribution of content. The algorithm should possess the following characteristics:

1) Adaptive adjustment to bandwidth changes: It should optimize the information quality of the ROI (ROI) in the image based on the available bandwidth.

2) Exploitation of spatio-temporal correlation: The algorithm should leverage the spatio-temporal correlation of content information in the video to improve efficiency.

To achieve this, we select the ROI area in the training dataset during the offline stage. We encode this area using the JPEG compression algorithm at different quality levels, measuring the time consumption and evaluating the inference accuracy of the encoded images. Subsequently, we generate a configuration set and query the library for the optimal configuration during online operations.

\subsection{Clustering Crops}
To optimize the information quality of the ROI in an image under a certain bandwidth, it is effective and intuitive to encode different areas of a video frame with different quality levels. This means preserving the high quality of pixels in the ROI while reducing the encoding precision of background areas. However, background information surrounding the target is also important, as deep learning models need to contrast the boundaries of foreground and background to extract target features, distinguish between them, and locate targets. The more boundary information there is, the better the extraction effect. At the same time, due to occlusion and the distribution of targets in the natural world, areas with dense target distributions are more likely to have new targets appearing.

\begin{figure}[h]
	\centering  
	\subfigbottomskip=2pt 
	\subfigcapskip=-5pt 
	\subfigure[Clustering num = 1]{
		\includegraphics[width=0.3\linewidth]{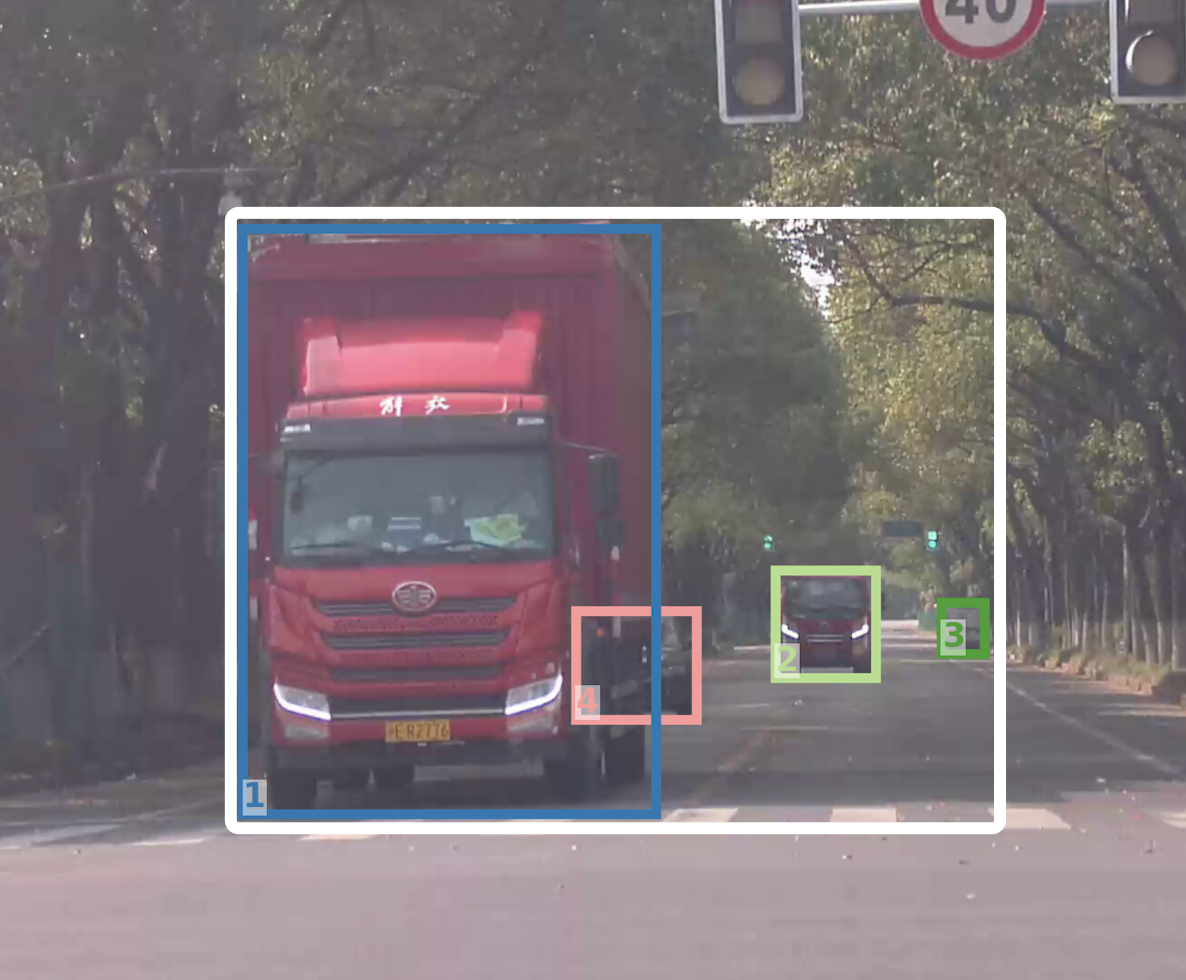}}
	\subfigure[Clustering num = 2]{
		\includegraphics[width=0.3\linewidth]{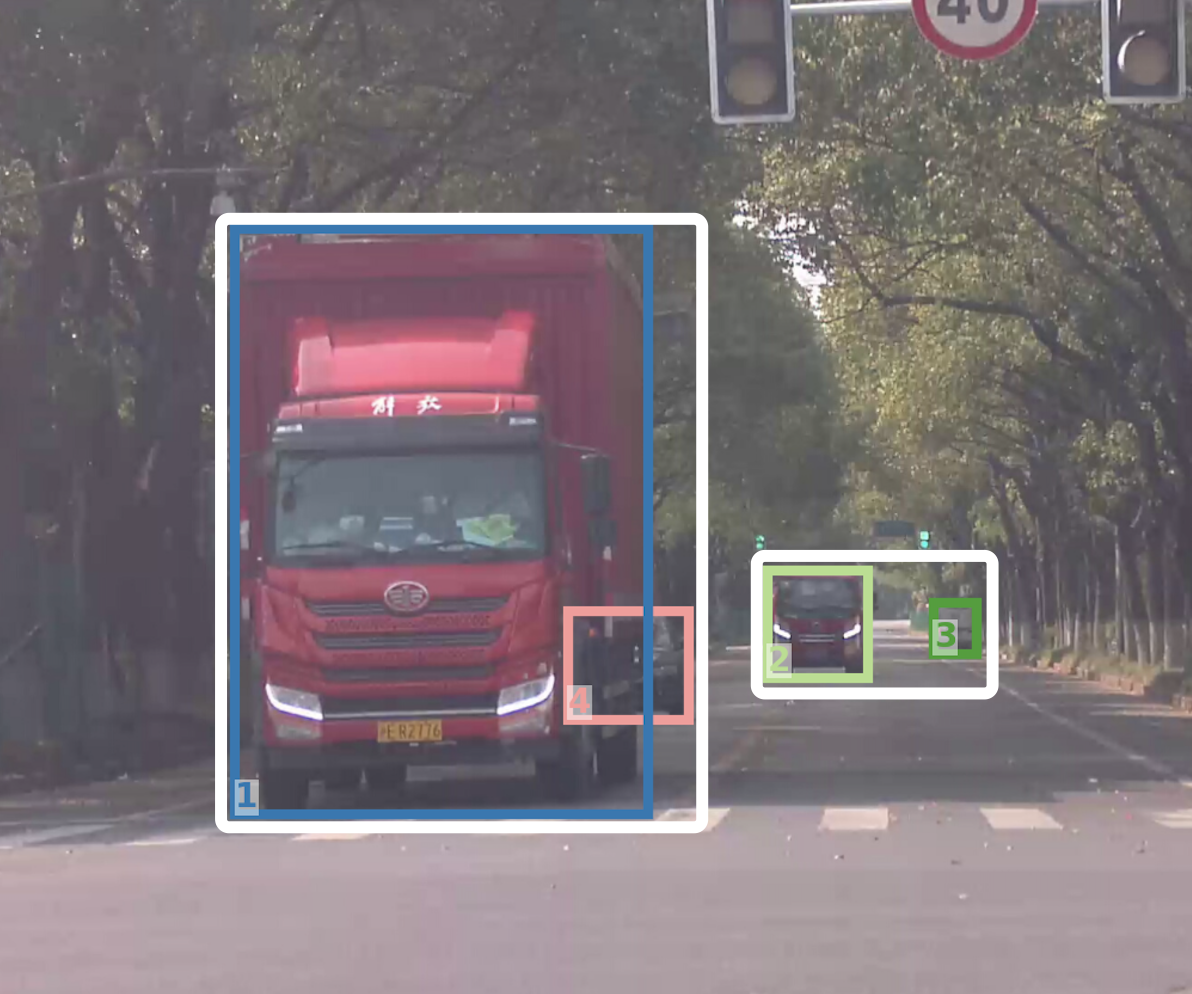}}
        \subfigure[Clustering num = 3]{
		\includegraphics[width=0.3\linewidth]{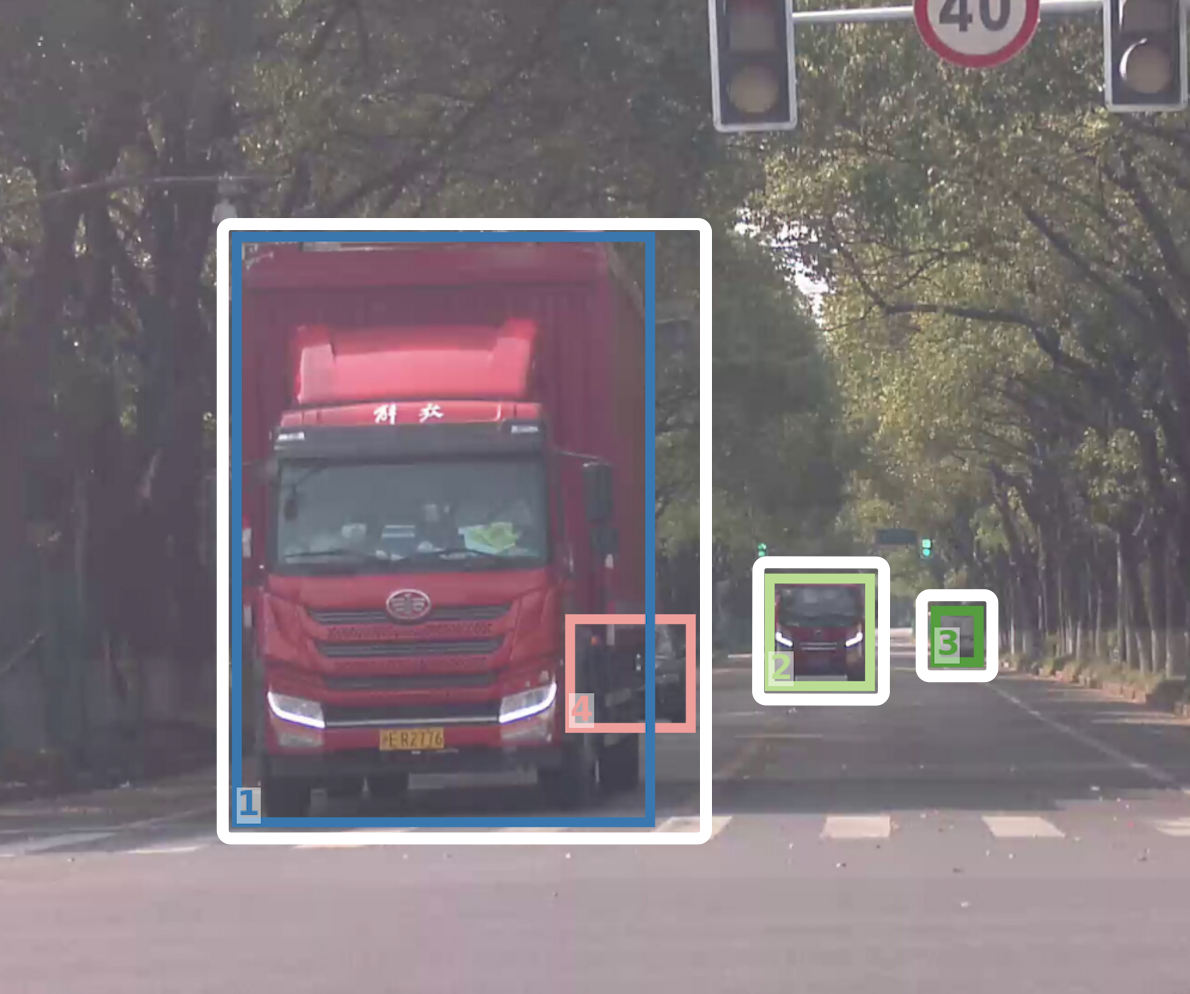}}
	\caption{Clustering crops: it shows ROI areas of the same frame under different cluster numbers. As the number of clusters increases, there are more discrete ROI areas with more accurate division.}
        \label{fig:crop}
\end{figure}

Therefore, it is necessary to preserve the background information around the ROI as much as possible. We achieve this goal by clustering targets according to their coordinates. Targets in the same cluster are cropped with a large rectangular box to form a single ROI. This ROI will include background information around the cluster of targets. During this process, the fewer clusters N, the larger the background area included in the cropped ROI. We hope that the clustering centers are closer to the center of the larger target areas in the cluster, making it easier for larger areas to be cropped separately and avoiding including too much background information. This way, the minimum bandwidth usage can be reduced to a lower level. Therefore, we use a weighted clustering algorithm with the size of the targets as weights, and the clustering results are shown in the Figure~\ref{fig:crop}. 


We also set the encoding quality coefficient Q dynamically according to the bandwidth. The clustering process will consume time and increase latency. We obtain the delay and file size of each video frame in the dataset under different configurations offline, and then upload them to the cloud for decoding, as shown in Algorithm \ref{alg:image_encoding}.

\begin{algorithm}
\DontPrintSemicolon
\caption{Quality Encode}
\SetAlgoLined
\label{alg:image_encoding}
\KwIn{Clustering number $K$, encode quality $Q$, detection boxes $\textbf{O}=\{O_1, O_2, ..., O_n\}$}
\KwOut{configuration set $S$}

\tcc{Step-1. BiKmeans clustering yields $K$ clusters $\textbf{C}=(C_1, C_2, ..., C_K)$}
initialize empty list $S$\\
compute the center point $t=(x,y)$ and weight $w$ for each detection box $O_i$\\
$\textbf{C} \leftarrow \operatorname{BiKmeans}(\underbrace{ t_{0},\cdot\cdot\cdot,t_{0} }_{w_{0}},\cdot\cdot\cdot, \underbrace{ t_{i},\cdot\cdot\cdot,t_{i} }_{w_{i}},\cdot\cdot\cdot)$ \\
\tcc{Step-2. Get $ROI$ and encode}
\For{$k=1$ to $K$}
{
$ROI_k \leftarrow \operatorname{Combine}(C_k)$ \\
}
$Bin \leftarrow \operatorname{Encode}(Frame|ROI,Q)$ \\
\tcc{Step-3. Get the accuracy}
$Img \leftarrow \operatorname{Decode}(Bin|ROI,Q)$,
$A=\operatorname{ServerModel.Inference}(Img)$ \\
\tcc{Step-4. Collect the result}
$S\left[\textbf{O}|K,Q \right]= A,Sizeof(Bin)$
\end{algorithm}

During the offline stage, we established a configuration set that uses the target distribution in the video frame, clustering number, and encoding quality as keys. We collected statistics on the accuracy and file size of the inference for the ROI area with differentiated encoding, as well as the time consumed for clustering and encoding. This information allows us to optimize the encoding process and select the best configuration for each situation during the online stage.

By using the \textit{Quality Encode Module}, we can adapt the encoding process according to the objects' importance and the content distribution in the video frames, ensuring efficient bandwidth usage and reducing latency. This approach allows us to better focus on the regions of interest and improve the overall performance of the object detection and tracking system.

\subsection{Configuration Set}
After generating configurations offline, the online clustering scheduler processes video frames sent to the cloud server during the online phase. It queries the configuration set based on current bandwidth and information such as target distribution in the video frame, and selects the optimal configuration based on the optimization objective. The target distribution in the video frame comes from the results of feature mining by the edge server. The key point of querying the configuration is to determine whether the target distribution in the current frame is consistent with that in the configuration set, which involves judging the target size, number, and similarity in relative positions. As the configuration set contains a large number of items, using the conventional method of calculating IOU would result in intolerable time loss. Moreover, IOU pays too much attention to the absolute position of target distribution, assuming that the ROI division after algorithm processing in step 6.2 should be the same when the target is shifted by the same magnitude in the image. However, at this point, IOU has undergone unpredictable changes. Therefore, we have established a target distribution similarity evaluation function with low computational complexity, which can embed the two-dimensional distribution of targets into a one-dimensional vector and convert the similarity of two-dimensional distributions into Mahalanobis distance of one-dimensional vectors. We divide the distribution into different regions, and each region has its own four-dimensional vector representing position and area to represent this two-dimensional distribution. However, it is possible that similar objects fall into different cells with a probability of one half, in which case we shift them two cells apart and add the distribution difference from two times. This fixes the difference in two-dimensional distributions to be reduced by 1.5 times, but it is still much smaller than the large differences. See the figure for details.

In this way, we obtain the vector representation of the configuration set. Since the one-dimensional vector is used, we use the product quantization algorithm to cluster the vector of the configuration set to speed up the query speed of the online clustering scheduler during online operation.

The optimization objective of the online clustering scheduler can be set to maximize the inference accuracy of cloud servers on each target $O_i$ in the video frame under the conditions of satisfying the bandwidth $B$ and the maximum delay $L$. See \myeqref{eq:config}

\begin{equation}
\begin{aligned}
&\max _{\left\{N,Q \right\}} \sum_{i} A_{O_{i}} \\
&C: \frac{\operatorname{Frame}_{ROI} \cdot Q_{ROI} + \operatorname{Frame}_{\sim ROI} \cdot Q_{\sim ROI}}{B}+t(K)+t(Q) \leqslant L
\end{aligned}
\label{eq:config}
\end{equation}

\noindent where $t(K)$ denotes the time cost of labeling and $t(Q)$ denotes the time cost of encoding. It is related to the number of clusters $K$ and the encoding quality $Q$, and the optimization objective can also be adjusted according to system requirements during operation.

\section{Evaluation}
\subsection{System Setup}
We implemented the various stages of our system using Python and adopted a camera-edge server-cloud server architecture. To accelerate inference speed, we deployed the deep learning models on the edge server using TensorRT. Communication between the edge server and cloud server was facilitated using the requests library. We validated the effectiveness of our system in complex network environments using an unmanned aerial vehicle, as shown in Figure \ref{fig:uav}, where the edge server was mounted on the UAV.

The hardware details of the platform, edge server, cloud server, and CNN models are as follows:

\begin{figure}
  \begin{minipage}{0.48\linewidth}
    \centering
    \includegraphics[width=0.97\linewidth]{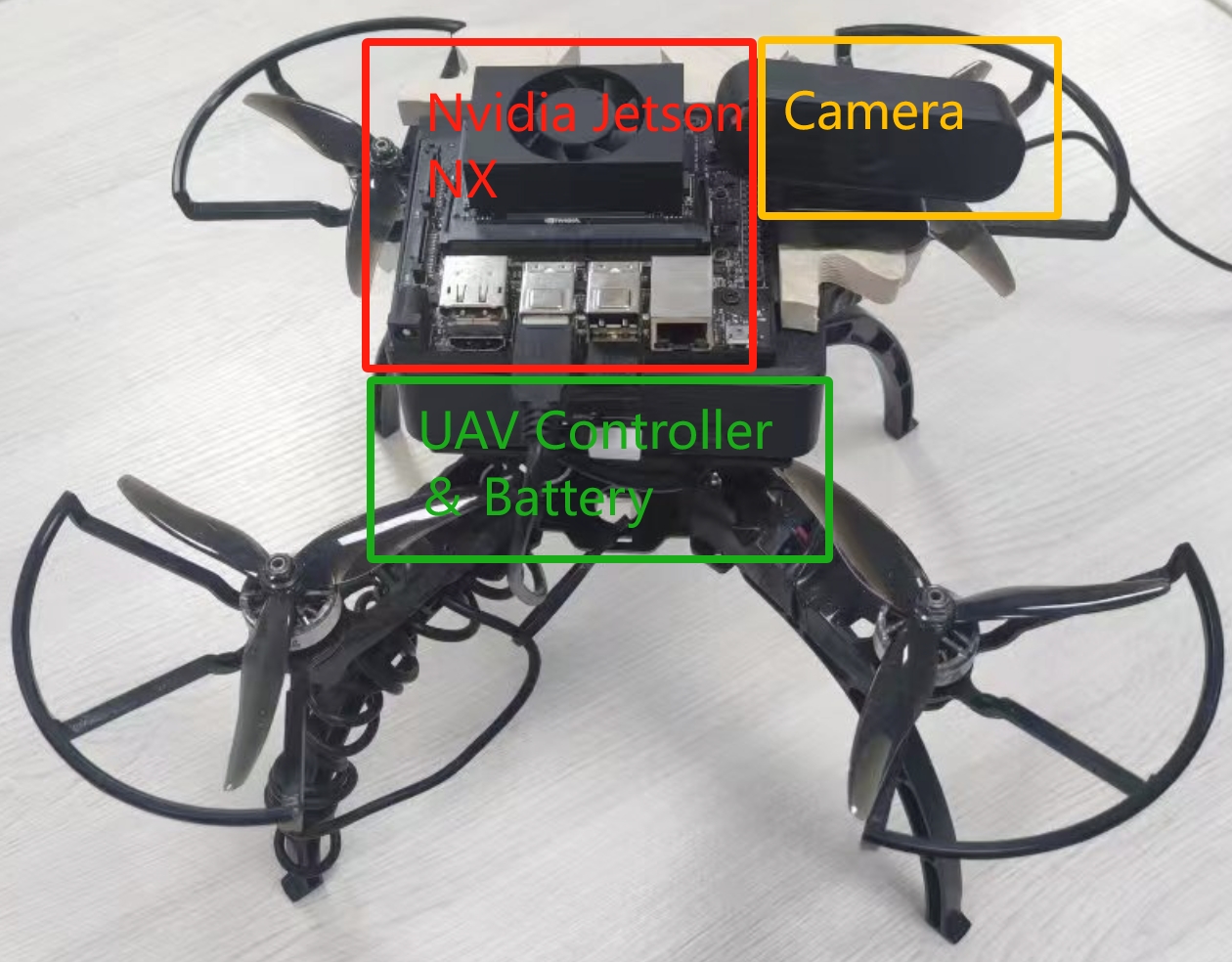}
    \caption{Platform: an UAV equipped with NVIDIA JETSON edge server and a HD camera.}
    \label{fig:uav}
  \end{minipage}%
  \hfill
  \begin{minipage}{0.48\linewidth}
    \centering
    \includegraphics[width=0.97\linewidth]{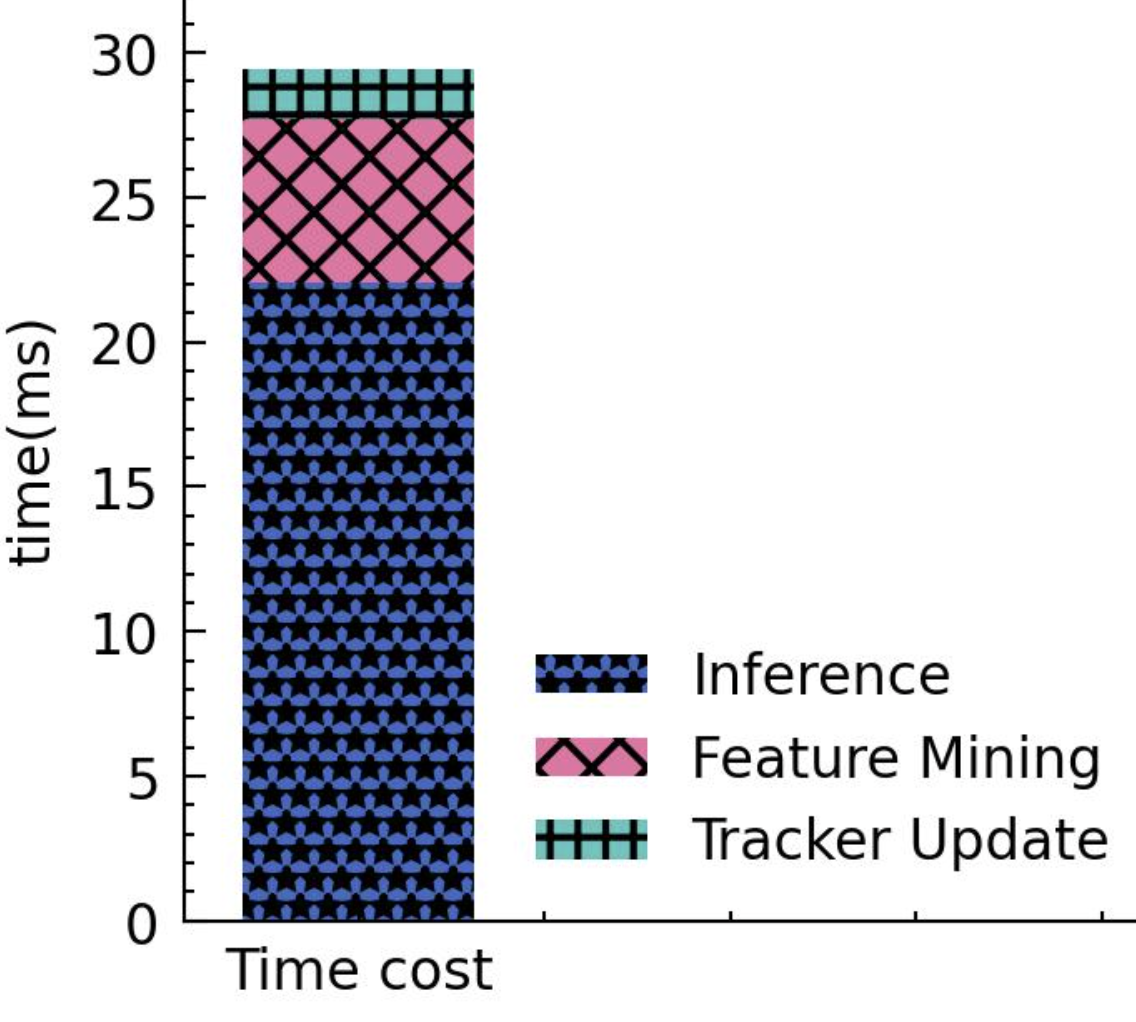}
    \caption{Time cost of each stage of the system working in the real scenario.}
     \label{fig:time cost}
  \end{minipage}
\end{figure}

\textbf{Platform:} Unmanned Aerial Vehicle

\textbf{Edge Server:} JETSON XAVIER NX (ARMv8 Processor ev0 V8l x6 cores, NVIDIA Tegra Xavier nvapu/integrated GPU, and 8GB memory)

\textbf{Cloud Server:} NVIDIA GTX 2080Ti x4

\textbf{CNN Models:} fasterrcnn\_mobilenet\_v3\_large\_fpn2\cite{howard2017mobilenets, ren2015faster} and fasterrcnn\_resnet18\cite{ren2015faster, he2016deep} on the edge server, yolo\_x\cite{ge2021yolox} on the cloud server

We conducted thorough performance evaluations of our system for object detection tasks in complex network environments. Our results show that with the support of the \textit{Fast Inference Module}, CloudEye reduces the inference time on the edge server by 24.55$\%$, achieving real-time video analysis at a processing speed of over 30 fps. Compared to performing inference solely on the edge server, CloudEye improved the accuracy by an average of 67.30$\%$. Moreover, CloudEye's dynamic differential quality encoding of video frames ensures a bandwidth reduction of 69.50$\%$ while maintaining accuracy.

\subsection{Latency}
In this section, we evaluate the latency level of the CloudEye system. During runtime, the majority of the frames are processed by the edge models combined with feature extraction to obtain the results. However, the key frames indicating significant changes in the content are encoded by the dynamic differential coding module and sent to the cloud for execution. 

First, we measure the actual performance of the system in the real scenario. We use the UAV to test the performance of CloudEye and the performance of traditional architecture systems under the same scenario and bandwidth conditions. 

\begin{figure}
    \begin{minipage}{0.48\linewidth}
    \centering
    \includegraphics[width=0.97\linewidth]{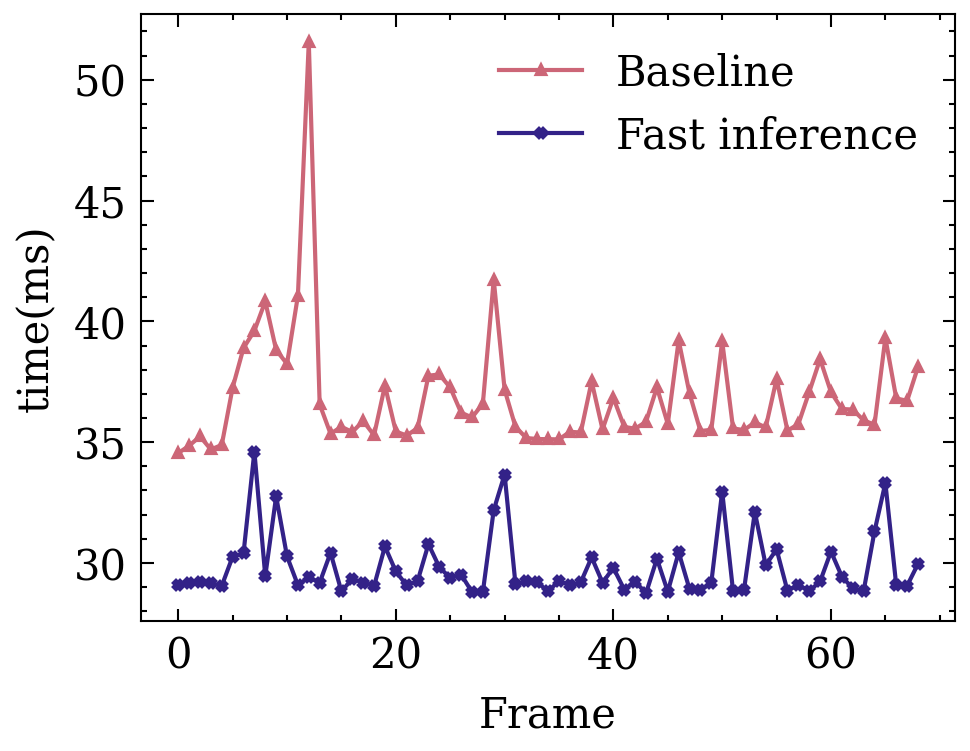}
    \caption{System latency in fast inference mode and normal mode as baseline in the real scenario.}
    \label{fig:fast latency}
  \end{minipage}%
  \hfill
  \begin{minipage}{0.48\linewidth}
    \centering
    \includegraphics[width=0.97\linewidth]{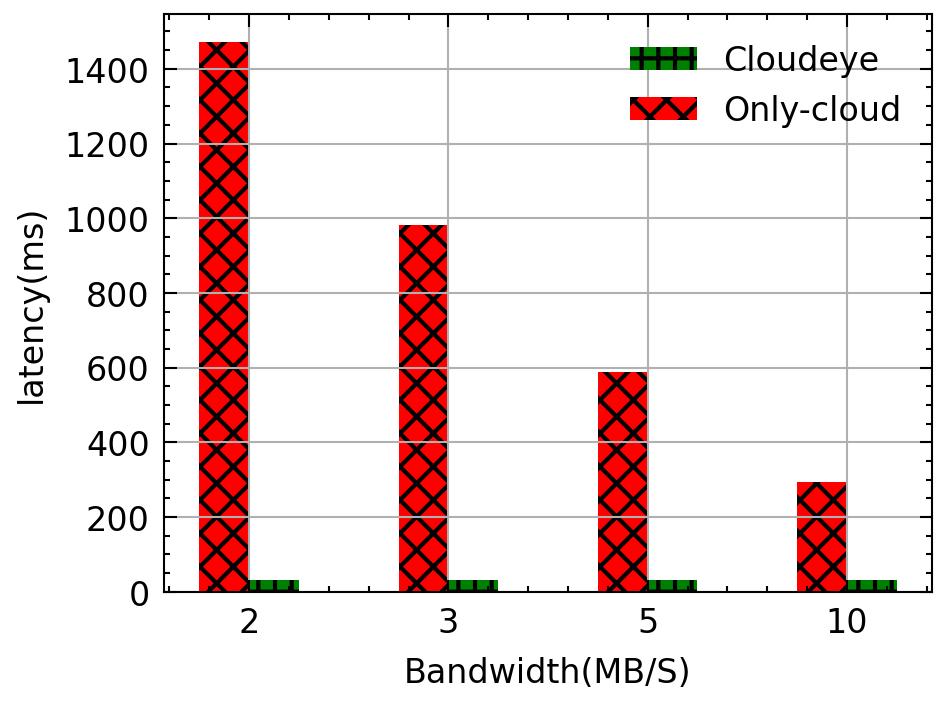}
    \caption{Comparison of time cost between Cloudeye and only-Cloud architecture system .}
    \label{fig:with cloud}
  \end{minipage}
\end{figure}

Figure ~\ref{fig:time cost} illustrates the primary components of the CloudEye's operational latency. It was observed that the primary delay is due to the edge model's inference, which is limited by the computation power of the edge server. Therefore, CloudEye employs the \textit{Fast Inference Module} to achieve real-time performance. To assess the impact of the \textit{Fast Inference Module} on inference time, we deploy the complete original model on the edge server for inference as a baseline. Figure ~\ref{fig:fast latency} shows the comparison of the time taken by the edge model under the two modes after removing the extreme values caused by the device's mechanical performance. The model under the fast inference mode and the baseline original model show the same fluctuation trend when inferring the same video frames. Evaluation results show that for the 2168*3848 resolution HD video, the average time for the original model converted by TensorRT to infer a single video frame with 16-bit floating-point precision is ~0.03683s. In contrast, under the fast inference mode, the average time for inferring a single video frame is ~0.02779s, which is 24.55$\%$ faster. This is mainly due to the guidance of the \textit{Fast Inference Module} on the model's proposal generation process. The \textit{Fast Inference Module} utilizes the spatio-temporal correlation of the video to provide the possible distribution region of the target as the ROI, which accelerates the model's proposal generation speed. Moreover, compared to the hundreds or thousands of proposals generated by the original model, the number of proposals generated by the \textit{Fast Inference Module} is more streamlined, making it faster and more direct for the model to regress the precise position of the target without having to handle numerous non-existent target proposals.

\begin{figure}
    \centering
    \includegraphics[width=0.97\linewidth]{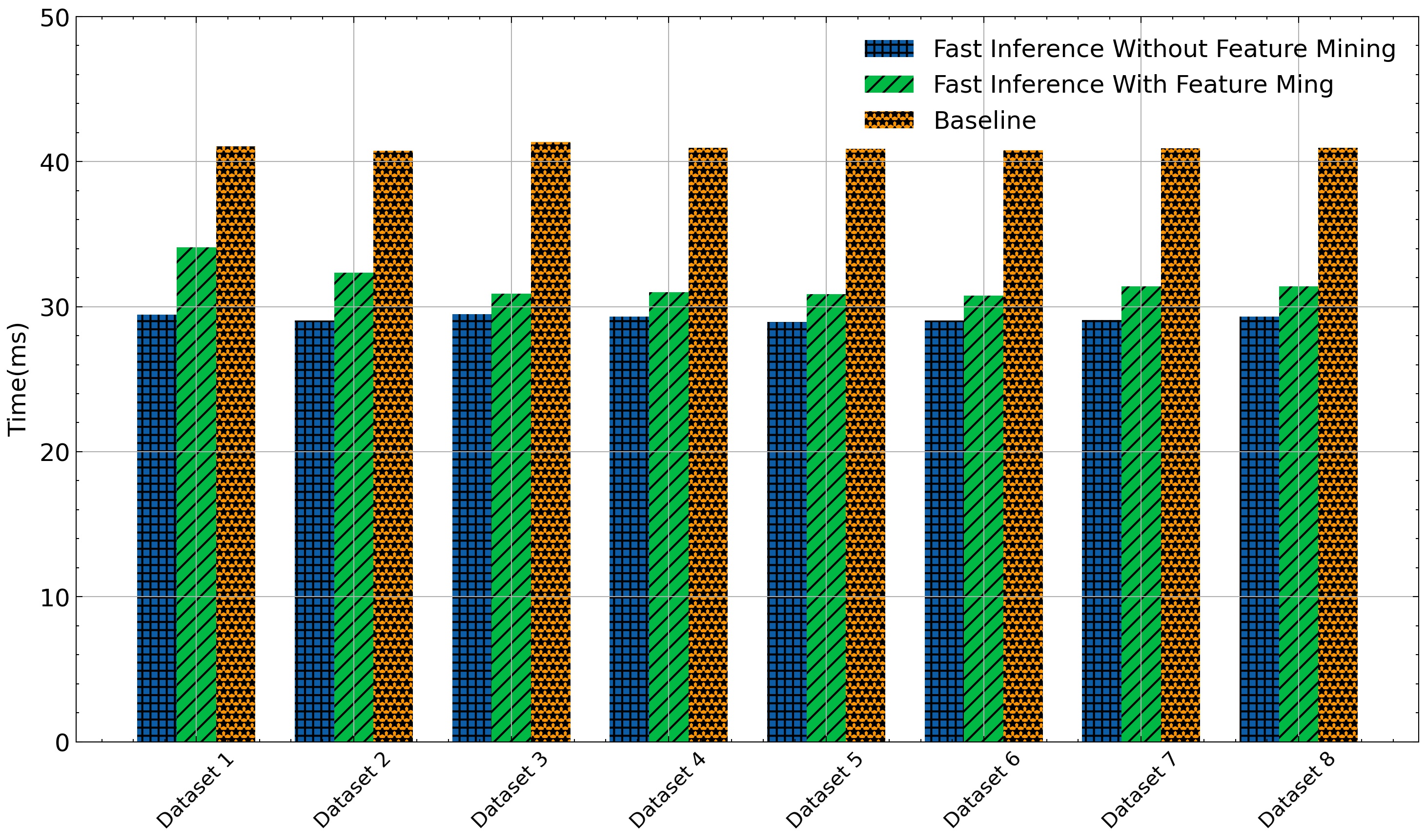}
    \caption{The latency of system in different modes.}
    \label{fig:with cloud1}
\end{figure}

Compared to the approach of uploading all frames to the cloud server for inference without employing edge servers, our system achieved significant performance advantages, as shown in Figure ~\ref{fig:with cloud1}. The approach of uploading all frames to the cloud server would result in intolerable latency and would be challenging to achieve real-time video analysis.

Besides, we also measure the system latency in different modes on the NUSCENES \cite{nuscenes} datasets. Figure ~\ref{fig:time cost} shows the latency of the system running in different modes on several scenes data sets. Here, the mode that system does not perform fast inference and feature mining is used as the baseline. The results show that \textit{Fast Inference Module} helps the system significantly reduce system latency, while \textit{Feature Mining Module} consumes very low latency.

\subsection{Accuracy}

\begin{figure}[h]
    \centering  
    \subfigbottomskip=2pt 
    \subfigcapskip=-5pt 
    \subfigure[Scenario A]{
        \includegraphics[width=0.48\linewidth]{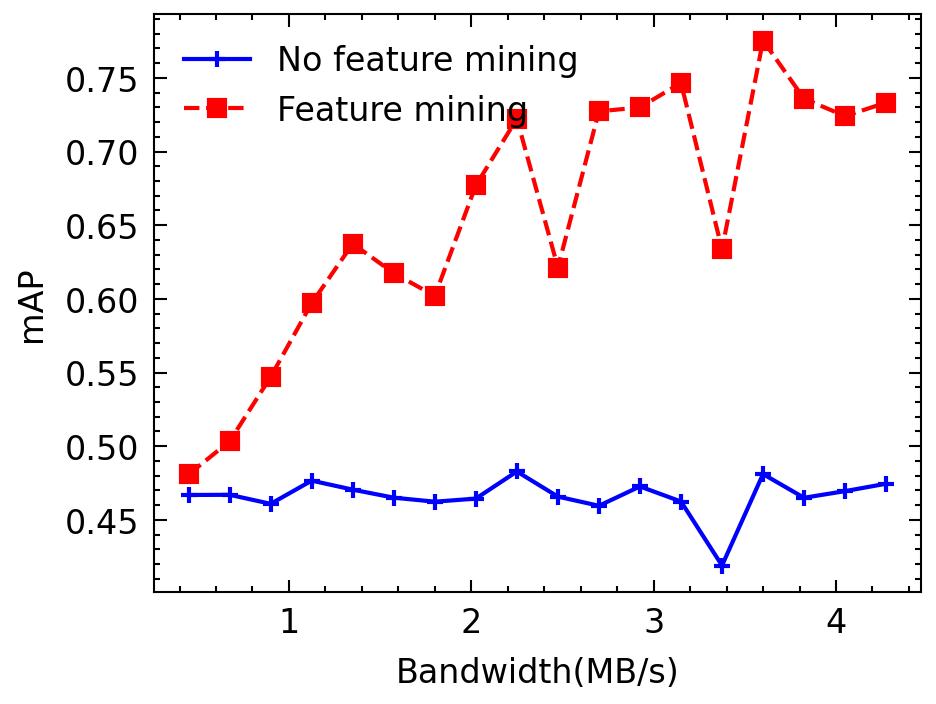}}
    \subfigure[Scenario B]{
        \includegraphics[width=0.48\linewidth]{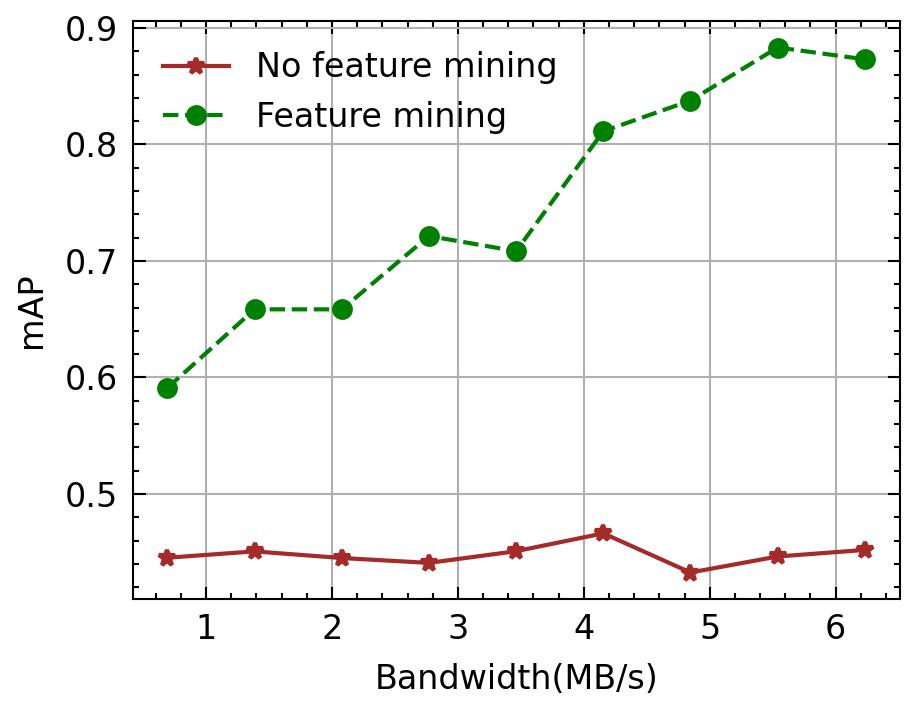}}
    \caption{System accuray (Mean Average Precision, mAP) of  feature mining mode and no feature mining mode in real scenarios.}
    \label{fig:real map}
\end{figure}

First, We measured the accuracy of CloudEye in the real scenario. Figure \ref{fig:real map} demonstrates the impact of the \textit{Fast Inference Module} of the edge model on accuracy. We use mAP as a metric to evaluate the average inference accuracy. According to the results, the model under the fast inference mode maintains comparable accuracy to that of the baseline model while reducing inference time. The average accuracy of the original model on the video dataset is about 0.4695, while the average accuracy of the model running under the fast inference mode is about 0.4223, which is only a limited accuracy loss while reducing inference time by 24.55$\%$.

However, this accuracy loss does not affect the system's efficiency, as the edge model's accuracy is inherently low due to limited computing power and resources of the edge server. As shown in Figure \ref{fig:mine}, the edge model mainly recognizes large and obvious objects and performs poorly on small and unclear ones. The role of the edge model is to infer the position of obvious objects in real-time. In realistic deployment scenarios, it does not need to independently recognize all targets. The \textit{Feature Mining Module} will track the targets that the edge model cannot recognize or recognizes incorrectly under the guidance of the cloud server. Finally, the average accuracy of the traditional edge-cloud architecture on the dataset is about 0.469, while the average accuracy of CloudEye is about 0.749, which represents a 67.30$\%$ accuracy improvement compared to the baseline. As shown in Figure \ref{fig:real map}, CloudEye has significant accuracy improvement compared to the traditional edge-cloud architecture in different bandwidth conditions. Even in poor bandwidth situations (<1MB/s), CloudEye can still achieve high accuracy. This also demonstrates the robustness of the feature mining algorithm, which can efficiently detect targets with low-frequency guidance from the cloud server. In good bandwidth conditions, CloudEye can obtain more accurate results from the cloud server, resulting in higher accuracy.

\begin{figure}
    \centering
    \includegraphics[width=0.97\linewidth]{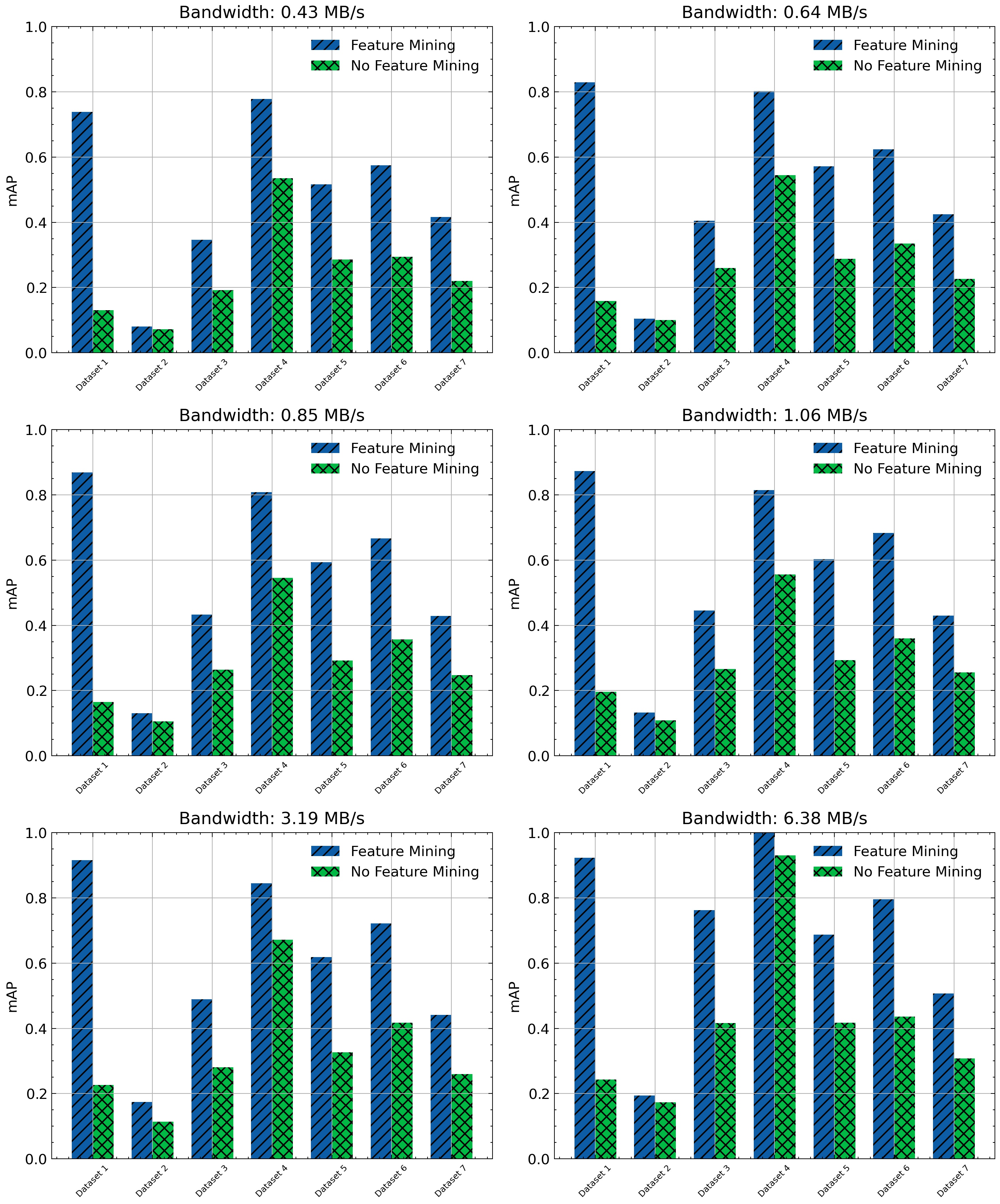}
    \caption{System accuray (Mean Average Precision, mAP) of  feature mining mode and no feature mining mode in NUSCENES \cite{nuscenes} datasets.}
    \label{fig:nuscenes map total}
\end{figure}

Besides, we also measure the system accuracy in different modes on the NUSCENES \cite{nuscenes} datasets with different bandwidth. As shown in Figure \ref{fig:nuscenes map total},  \textit{Feature Mining Module} enables CloudEye to achieve higher accuracy in different data sets. As bandwidth increases, ClouEye's accuracy increases. However, CloudEye still performs well under low bandwidth conditions. 

\begin{figure}
    \begin{minipage}{0.48\linewidth}
    \centering
    \includegraphics[width=0.97\linewidth]{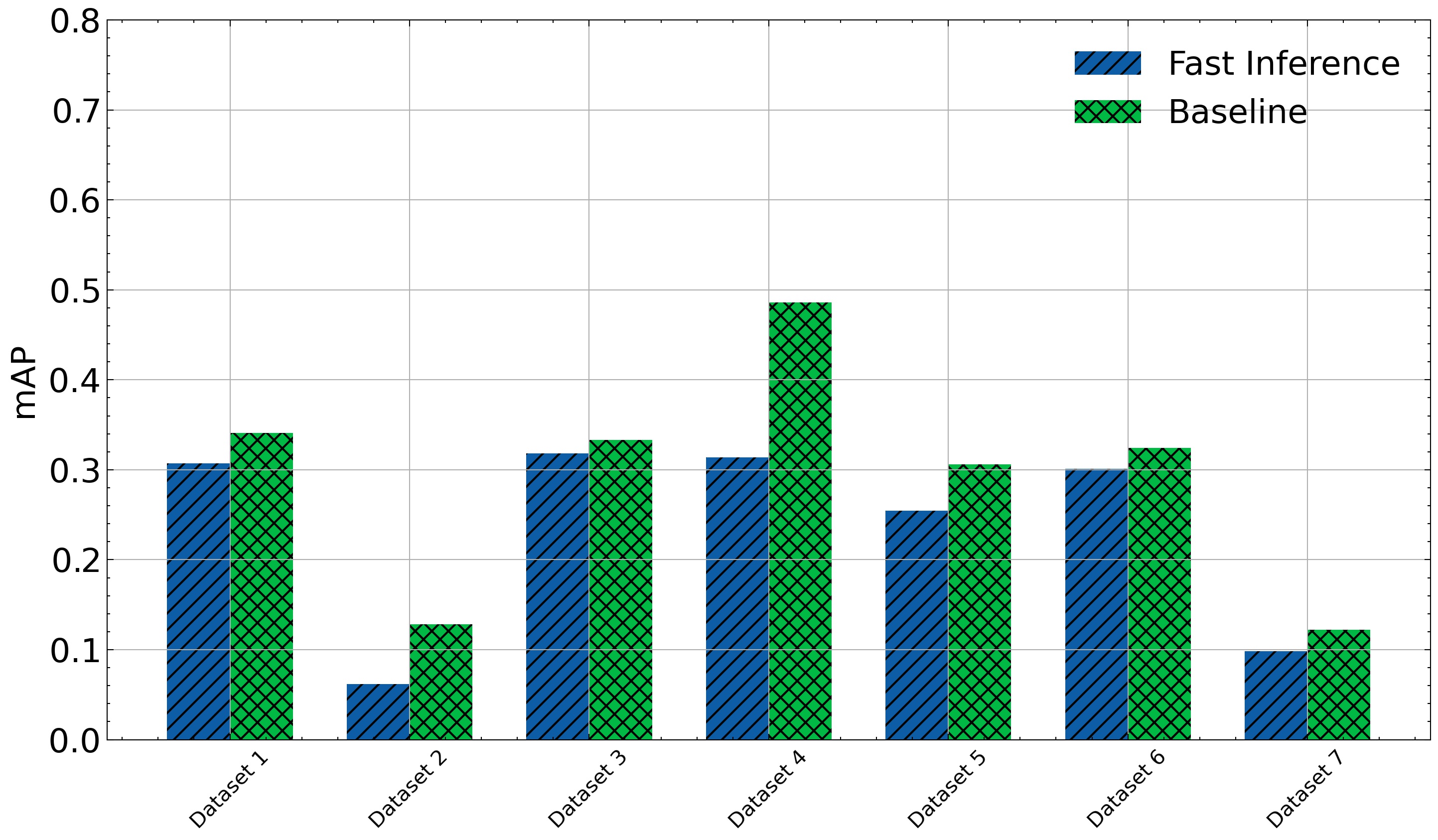}
    \caption{System accuray with resnet18\cite{he2016deep} on edge server.}
    \label{fig:acc 1}
  \end{minipage}%
  \hfill
  \begin{minipage}{0.48\linewidth}
    \centering
    \includegraphics[width=0.97\linewidth]{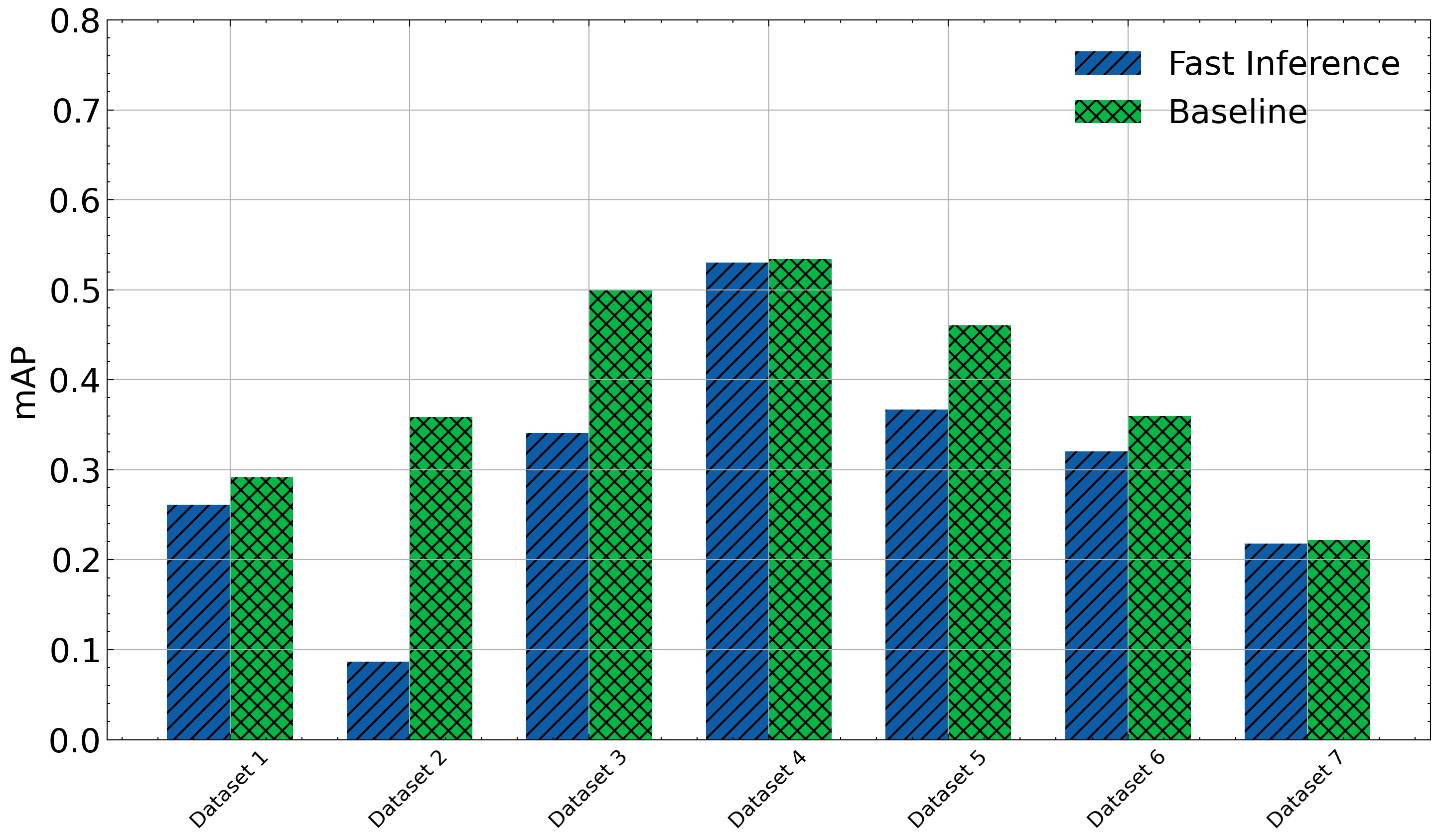}
    \caption{System accuray with mobilenet\cite{howard2017mobilenets} on edge server.}
    \label{fig:acc 2}
  \end{minipage}
\end{figure}

At the same time, we test the accuracy performance of different edge models in feature mining mode. Figure \ref{fig:acc 1} and \ref{fig:acc 2}shows that \textit{Feature Mining Module} can significantly improve the accuracy of models of different architectures.

\subsection{Bandwidth}

\begin{figure}
  \begin{minipage}{0.48\linewidth}
    \centering
    \includegraphics[width=0.97\linewidth]{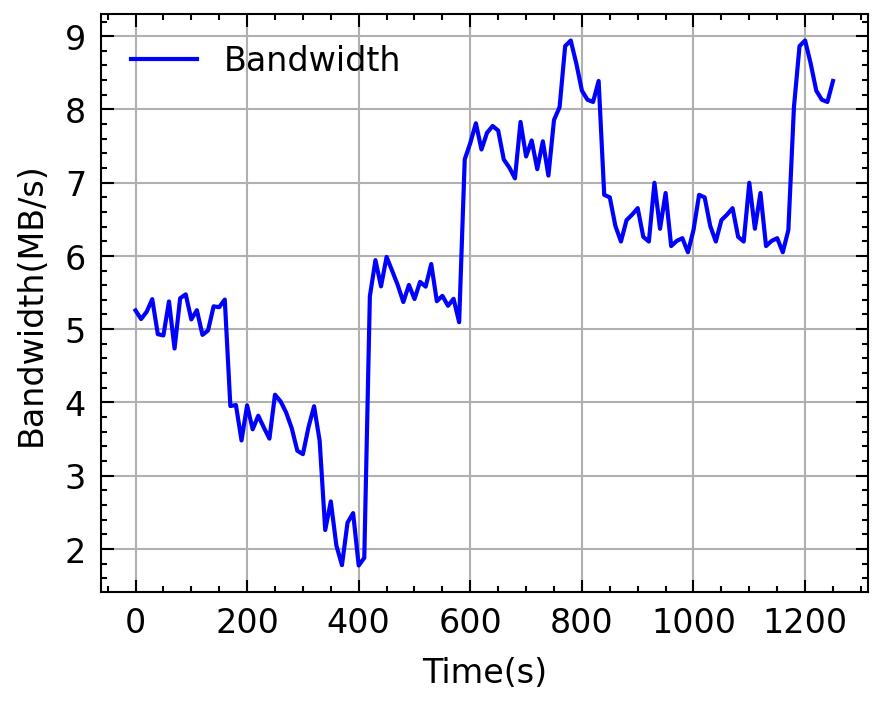}
    \caption{Bandwidth between the edge server and the cloud server}
    \label{fig:band}
    \Description{}
  \end{minipage}%
  \hfill
  \begin{minipage}{0.48\linewidth}
    \centering
    \includegraphics[width=0.97\linewidth]{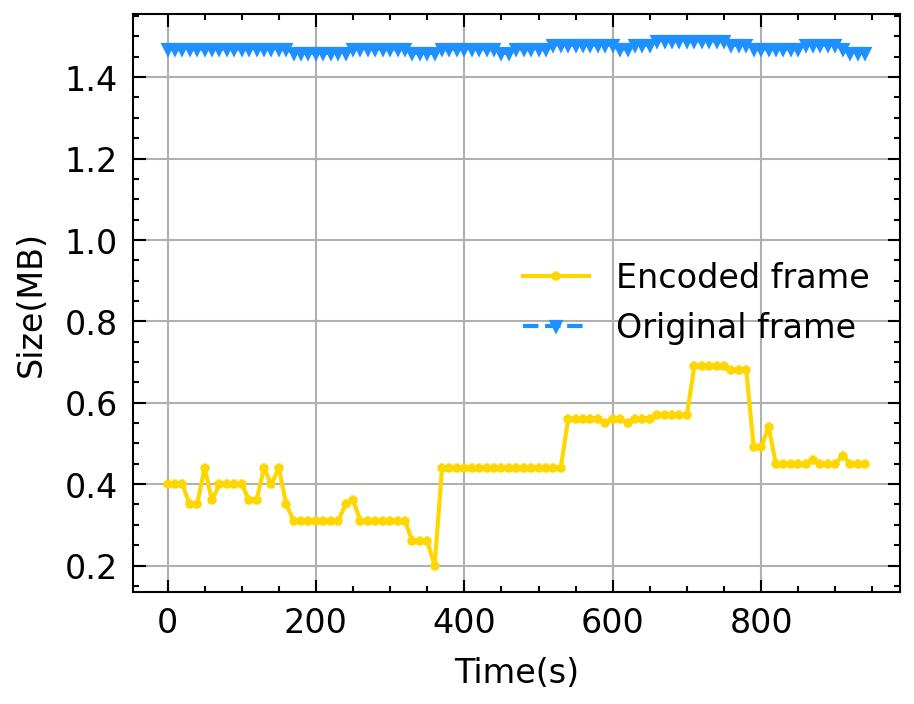}
    \caption{The size of the encoded video frames after quality encoding and original video frames}
    \label{fig:depress}
    \end{minipage}
\end{figure}

In this section, we will evaluate the bandwidth of the system during its actual operation. In the real-world test with the UAV carrying the edge server, we obtained the bandwidth between the edge server and the cloud server as shown in Figure~\ref{fig:band}. Under this condition, we obtained the size of the encoded frames after differential quality encoding of the video frames. As shown in Figure~\ref{fig:depress}, the average size of the original video frames is around 1.472MB. 

According to the bandwidth changes and video content differences, the system clustered the target areas based on the distribution of the targets in the video, and the dynamic encoding module performed differential quality encoding on the regions of interest and non-interest. The resulting average size of the encoded frames was around 0.448MB, which means that the system saved 69.50$\%$ of the bandwidth while ensuring the encoding quality of the regions of interest.

\section{CONCLUSION AND FUTURE WORK}

In conclusion, we have designed and implemented CloudEye, a lightweight and efficient edge-cloud hybrid architecture for video analysis in mobile visual scenes. Our contribution lies in the exploration and research of spatio-temporal correlation in video analysis tasks, as well as the establishment of a coupled optimization mechanism based on the interested regions throughout the entire process. We have also uncovered a vast optimization space in the deep learning model at a lower level, such as the model's microscopic structure and operator level, to minimize the system's computational and resource consumption. In addition to the significant contributions made in our research, the CloudEye is highly adaptable and easily deployable in a wide range of applications, allowing for on-demand customization of deep learning models to address specific video analysis needs. With its modular design and strong scalability, CloudEye can be used as a plug-and-play solution for a variety of video analysis systems. Future research can focus on expanding and optimizing the system's modules, developing new models, and integrating with other systems to achieve even greater functionality and efficiency.

Moving forward, our research will focus on exploring how to leverage the deep learning features of the edge model for more robust modeling and tracking of targets, to achieve better mining accuracy and efficiency. Moreover, we will further integrate the various modules of the system and study the relationship between deep learning features and differentiated quality coding based on interested regions, to maximize the system's efficiency.

\bibliographystyle{ACM-Reference-Format}
\bibliography{refs}
\end{document}